 \documentclass[journal]{IEEEtran}
\usepackage{epsfig}
\usepackage{graphicx}
\usepackage{amsmath}
\usepackage{amssymb}
\usepackage{amsmath} 
\usepackage{amsmath} 
\usepackage{algorithmic}
\usepackage{algorithm}
\usepackage{subfigure}
\usepackage{multirow}
\usepackage{url}
\usepackage{color}
\usepackage{hyperref}
\usepackage{balance}
\usepackage{setspace}
\usepackage{slashbox}
\usepackage{booktabs}
\usepackage{threeparttable}

\renewcommand{\baselinestretch}{1.0}

\ifCLASSOPTIONcompsoc
  \usepackage[nocompress]{cite}
\else
  \usepackage{cite}
\fi

\ifCLASSINFOpdf
\else
\fi

\begin{document}
\title{Crossbar-Net: A Novel Convolutional Neural Network for Kidney Tumor Segmentation in CT Images}

\author{Qian~Yu, Yinghuan~Shi$^*$, Jinquan~Sun, Yang~Gao$^*$, Jianbing~Zhu, Yakang~Dai% <-this % stops a space
\IEEEcompsocitemizethanks{
%\IEEEcompsocthanksitem This work was supported by the NSFC (61432008, 61673203), Young Elite Scientists Sponsorship Program by CAST (YESS 2016QNRC001), CCF-Tencent Open Research Fund (RAGR 20180114), Projects of Shandong Province Higher Educational Science and Technology Program (J18KA370, J15LN58), Project of Shandong Medicine and Health Science Technology Development Plan (2017WSB04071), and Shandong Province Science and Technology Development Plan Project (2014GSF118086).
%This work was supported by Zhejiang Key Technology Research Development Program (2018C03024), Jiangsu Key Technology Research Development Program (BE2017664), Suzhou Science and Technology Projects for People's Livelihood (SYS2018010), Suzhou Science and Technology Development Project (SZS201818), and SND Medical Plan Project (2016Z010, 2017Z005).
%\protect\\
\IEEEcompsocthanksitem \emph{Corresponding Authors: Yang Gao and Yinghuan Shi}.
\IEEEcompsocthanksitem Qian Yu, Yinghuan Shi, Jinquan Sun, and Yang Gao are with the State Key Laboratory for Novel Software Technology, the National Research Institute for Big Data Science in Health and Medicine, and the Collaborative Innovation Center of Novel Software Technology and Industrialization, Nanjing University, China.
Qian Yu is also with School of Data and Computer Science, Shandong Women's University, China. (e-mail: yuqian@sdwu.edu.cn, syh@nju.edu.cn, jinquansun@gmail.com, gaoy@nju.edu.cn)%\protect\\
\IEEEcompsocthanksitem Jianbing Zhu is with the Suzhou Science and Technology Town Hospital, China. (e-mail: zeno1839@126.com)%\protect\\
\IEEEcompsocthanksitem Yakang Dai is with the Suzhou Institute of Biomedical Engineering and Technology, Chinese Academy of Sciences, China. (e-mail: daiyk@sibet.ac.cn)%\protect\\
}% <-this % stops an unwanted space
}

%\markboth{IEEE Transactions on Image Processing}%
%{Shi \MakeLowercase{\textit{et al.}}}



\IEEEtitleabstractindextext{%
\begin{abstract}
Due to the unpredictable location, fuzzy texture and diverse shape, accurate segmentation of the kidney tumor in CT images is an important yet challenging task. To this end, we in this paper present a cascaded trainable segmentation model termed as Crossbar-Net. Our method combines two novel schemes: (1) we originally proposed the crossbar patches, which consists of two orthogonal non-squared patches (\emph{i.e.}, the vertical patch and horizontal patch). The crossbar patches are able to capture both the global and local appearance information of the kidney tumors from both the vertical and horizontal directions simultaneously. (2) With the obtained crossbar patches, we iteratively train two sub-models (\emph{i.e.}, horizontal sub-model and vertical sub-model) in a cascaded training manner. During the training, the trained sub-models are encouraged to become more focus on the difficult parts of the tumor automatically (\emph{i.e.}, mis-segmented regions). Specifically, the vertical (horizontal) sub-model is required to help segment the mis-segmented regions for the horizontal (vertical) sub-model. Thus, the two sub-models could complement each other to achieve the self-improvement until convergence. In the experiment, we evaluate our method on a real CT kidney tumor dataset which is collected from 94 different patients including 3,500 CT slices. Compared with the state-of-the-art segmentation methods, the results demonstrate the superior performance of our method on the Dice similarity coefficient, true positive fraction, centroid distance and Hausdorff distance. Moreover, to exploit the generalization to other segmentation tasks, we also extend our Crossbar-Net to two related segmentation tasks: (1) cardiac segmentation in MR images and (2) breast mass segmentation in X-ray images, showing the promising results for these two tasks. Our implementation is released at \url{https://github.com/Qianyu1226/Crossbar-Net}.

\end{abstract}

% Note that keywords are not normally used for peer review papers.
}

\maketitle
\IEEEdisplaynontitleabstractindextext
\IEEEpeerreviewmaketitle
\bigskip
\bigskip

\IEEEraisesectionheading{\section{Introduction}}
\label{sec:introduction}
%figure 1
\begin{figure}[tb]
\centering
\setlength{\belowcaptionskip}{-1cm}
\subfigure[]{
 \label{tumors_examples:a}
 \includegraphics[width=0.9in, height=0.9in]{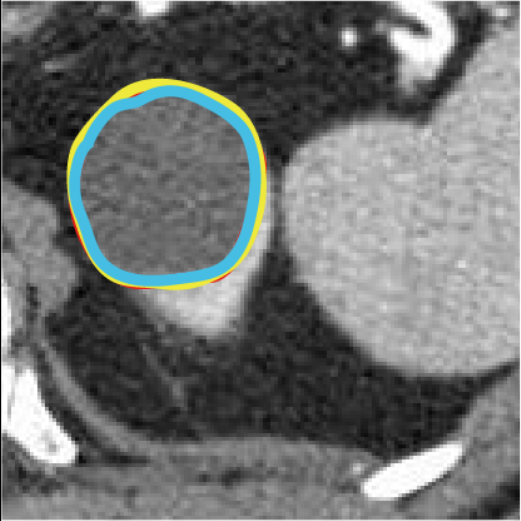}
}
\subfigure[]{
 \label{tumors_examples:b}
 \includegraphics[width=0.9in, height=0.9in]{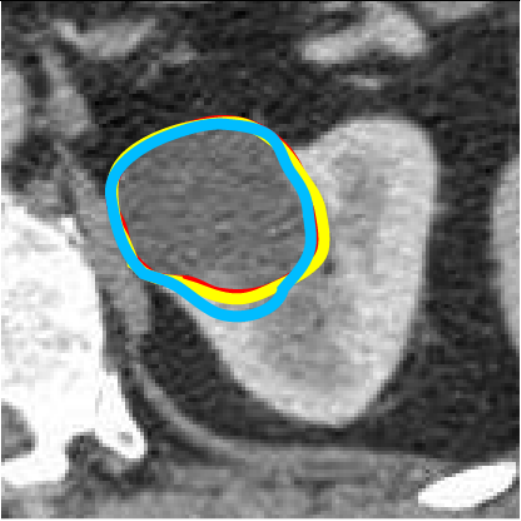}
}
\subfigure[]{
 \label{tumors_examples:c}
 \includegraphics[width=0.9in, height=0.9in]{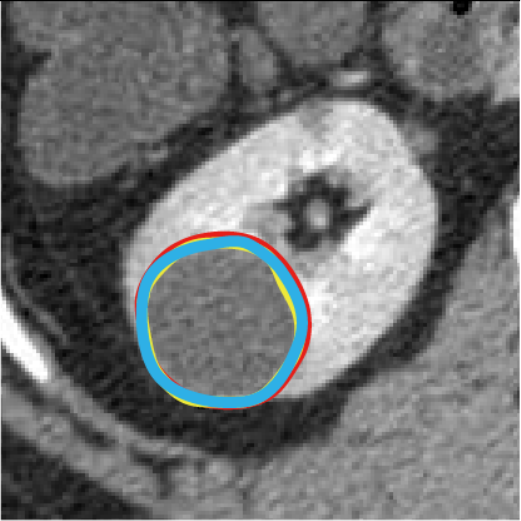}
}
\subfigure[]{
 \label{tumors_examples:d}
 \includegraphics[width=0.9in, height=0.9in]{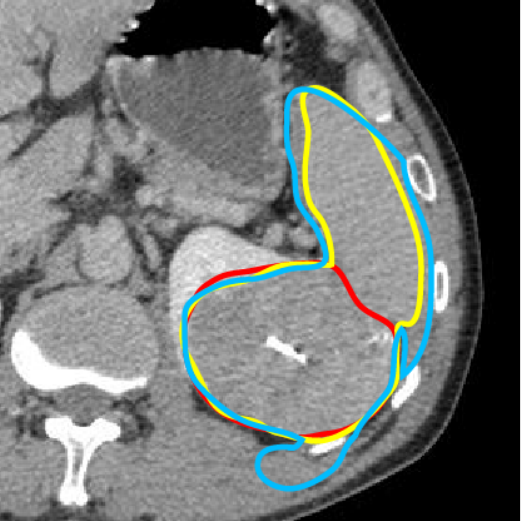}
}
\subfigure[]{
 \label{tumors_examples:e}
 \includegraphics[width=0.9in, height=0.9in]{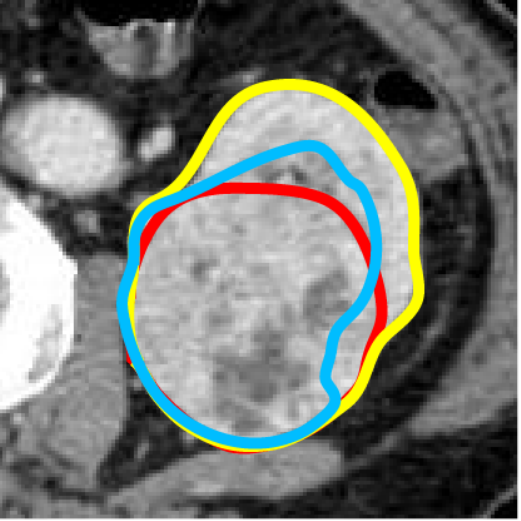}
}
\subfigure[]{
 \label{tumors_examples:f}
 \includegraphics[width=0.9in, height=0.9in]{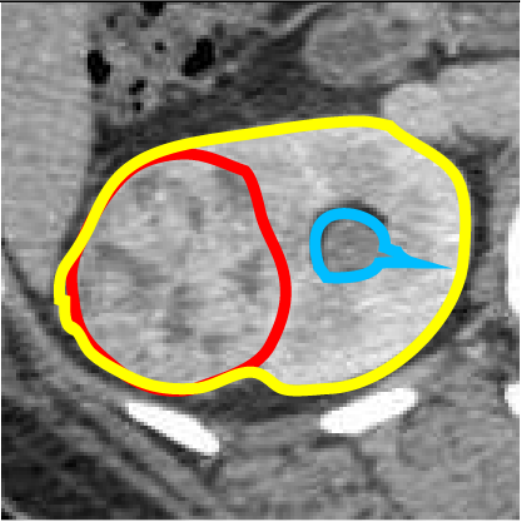}
}
\subfigure[]{
 \label{tumors_examples:g}
 \includegraphics[width=0.9in, height=0.9in]{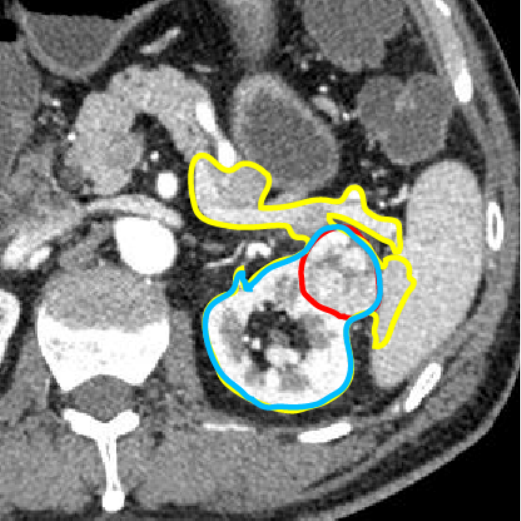}
}
\subfigure[]{
 \label{tumors_examples:h}
 \includegraphics[width=0.9in, height=0.9in]{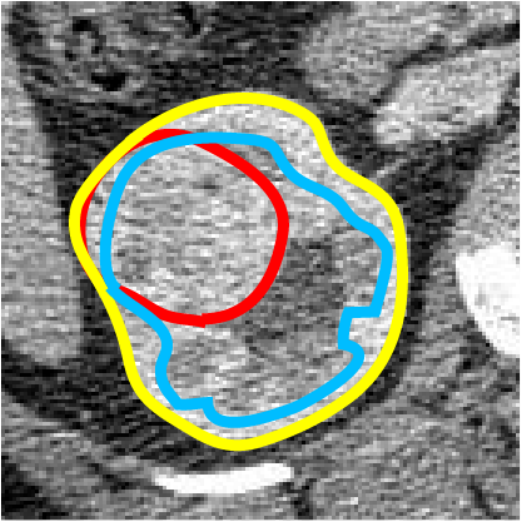}
}
\subfigure[]{
 \label{tumors_examples:i}
 \includegraphics[width=0.9in, height=0.9in]{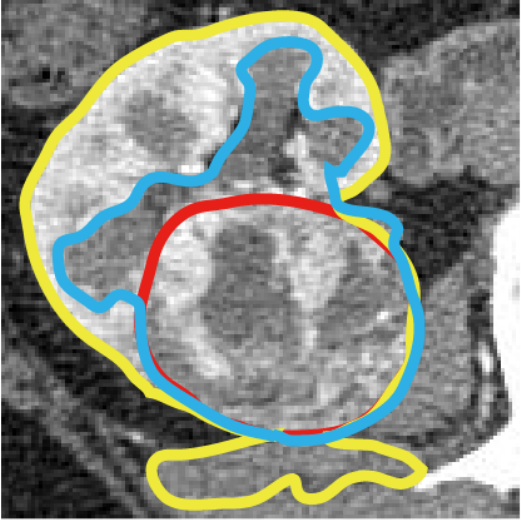}
}
\caption{Typical images of kidney tumors. The red, yellow and light blue contours denote the ground truth, predictions of the energy-based method \cite{linguraru2011automated} and traditional learning-based method \cite{skalski2016kidney}, respectively. Extensive comparisons with other state-of-the-art models are reported in Section \ref{sec:experiments}.}\label{tumors_examples}
\end{figure}
\IEEEPARstart{R}{enal} cell carcinoma is a common urologic cancer arising from renal cortex \cite{oh2018accurate, facts2018cancer, xiang2017cortexpert:}. Accurate quantification and correct classification of tumors could largely influence the effect of the following computer-aided treatment of renal cell carcinoma \cite{kim2004computer}. In this meaning, for the quantification and classification, the accurate kidney tumor segmentation is a significant prerequisite. Traditional human-based manual delineation for kidney tumor segmentation is not desirable in clinical practice, due to both the subjective (\emph{e.g.}, incorrect delineation) and objective (\emph{e.g.}, a large number of images) factors. Thus, computer-aided automatic segmentation methods for kidney tumors (in CT images) are in high demand.

However, segmenting the kidney tumors automatically in CT images is a very challenging task. According to the clinical and experimental observation,
\begin{itemize}
\item The location of different kidney tumors in medical images is difficult to predict since the tumors could possibly appear in very different places between different patients.
\item Different tumors between different patients usually show very diverse shape appearance and volumetric size according to the different growth stages.
\item The tumors and their surrounding tissues are with very similar texture information due to the low contrast of CT images.
\end{itemize}
Although several works have been proposed recently \cite{kim2004computer, lee2017detection, hodgdon2015can, linguraru2009computer, skalski2016kidney, linguraru2011automated}, their segmentation performance could not be robustly guaranteed in different cases (see Fig. \ref{tumors_examples}). Both (1) the intensity dissimilarity within different parts of tumors and (2) the similar appearance between kidney tumors and their surrounding tissues pose the great technical challenges for developing robust segmentation models. In order to visually illustrate these challenges, we have segmented several typical kidney tumors in CT images by introducing two representative models: the energy minimization-based model \cite{linguraru2011automated} and the traditional learning-based model \cite{skalski2016kidney}.  Please note that the extensive comparison with other state-of-the-art models is reported in our experimental part. As shown in Fig. \ref{tumors_examples:a} - \ref{tumors_examples:c}, the tumors with high contrast and clear boundaries could be well segmented by traditional segmentation methods \cite{linguraru2011automated, skalski2016kidney}. However, \cite{linguraru2011automated, skalski2016kidney} will fail in more difficult cases. For example, the tumor in Fig. \ref{tumors_examples:d} is strongly connected to its surrounding tissue and meanwhile shows a similar intensity with that of the tissue, which leads \cite{linguraru2011automated, skalski2016kidney} fail to segment. Similarly, all the tumors in Fig. \ref{tumors_examples:e} - Fig. \ref{tumors_examples:f} show very similar visual characteristics with the outside renal parenchyma. In addition, the tumors in Fig. \ref{tumors_examples:g} - Fig. \ref{tumors_examples:i} are challenging cases because they are with intensity dissimilarity within different parts inside the tumor. Therefore, the advanced efforts on accurate kidney tumor segmentation are still required to meet the clinic requirement.

The key issue of accurate segmentation is how to well distinguish the tumor and non-tumor boundary by extracting (or learning) the informative and discriminative features. Recent trends of deep convolutional neural network (CNN) have demonstrated the superior performance on learning-based segmentation tasks in different imaging modalities for different organs, \emph{e.g.}, prostate \cite{shi2017does, he2018automatic}, heart \cite{mortazi2017cardiacnet:,khened2018densely, patravali20172d-3d}, brain \cite{moeskops2016automatic, wang2018interactive, zhang2018automatic}. Hence, in this paper, we present a CNN-based model for CT kidney tumor segmentation. Previous CNN-based segmentation methods could be roughly classified into two categories: the image-based CNN models \cite{long2015fully, ronneberger2015u, milletari2016v, zhang2017combining, lalonde2018capsules} and the patch-based CNN models \cite{ciresan2012deep, wang2017central, shi2017does}. Both of these previous methods treated either whole images or squared patches as the training samples to first learn the segmentation model and then employ the obtained models to segment the new coming testing images.

%figure crossbar patch
\begin{figure}[tb]
\setlength{\belowcaptionskip}{-0.5cm}
\centering
\subfigure[]{
 \label{single_example:a}
 \includegraphics[width=0.9in, height=0.9in]{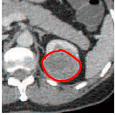}
}
\hspace{2pt}
\subfigure[]{
 \label{single_example:b}
 \includegraphics[width=0.9in, height=0.9in]{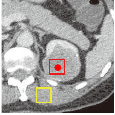}
}
\hspace{2pt}
\subfigure[]{
 \label{single_example:c}
 \includegraphics[width=0.9in, height=0.9in]{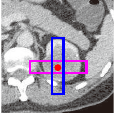}
}
\caption{An example of the kidney tumors and different types of patches. (a) Ground truth. (b) Squared patches. (c) Our crossbar patches.}\label{fig:single-example}
\end{figure}

Unlike the traditional image- or patch-based CNN models, we originally propose the new findings for kidney tumor segmentation in CT images. Specifically, in CT images, the kidney tumors normally appear as the subrounded shape with a certain degree of symmetry. This observation inspires us that we could leverage the shape information for kidney tumor segmentation to achieve a promising performance. Unfortunately, this specific shape information is usually ignored in previous CNN-based methods. For example, if we extract squared patches as the existing patch-based CNN models, as shown in Fig. \ref{single_example:b}, the red and yellow patches with only captured local information cannot distinguish the tumor and its surrounding organs obviously. Alternatively, we can enlarge the size of the squared patches or apply the whole image-based methods directly to cover the whole tumor and its context. However, in this case, the irrelevant noise might be brought in at the same time, which might cause that the local details, especially the boundary details, might be ignored. The following results in Fig. \ref{dr_and_tpf} and Table \ref{table_adnc} support this point. Basically, it is known that, in a same region area, compared with the squared patch, the non-squared rectangular patch could capture more information typically from one direction (\emph{i.e.}, horizontal or vertical). If we sample non-squared patches (named as crossbar patches in this paper) as Fig. \ref{single_example:c} to fully cover the whole tumor along one direction from side-to-side, we indeed could integrate more contextual and symmetrical information simultaneously.

Thus, we innovatively in this paper propose crossbar patches which consist of the vertical patch and horizontal patch, aiming to jointly capture (1) the local detail information and (2) global contextual information from vertical and horizontal directions, respectively. In addition, on the obtained crossbar patches, we originally present a cascaded training framework to iteratively train the sub-models (namely vertical sub-model and horizontal sub-model) from these two directions. It is noteworthy that our training and testing are performed on the pixel-wise since we convert the segmentation task to a pixel-wise classification problem as the traditional setting \cite{ciresan2012deep, wang2017central, shi2017does}.

In particular, during the training process, the trained vertical and horizontal sub-models are encouraged to help each other in a way of asking the other one to help segment its difficult parts. Taking the vertical sub-model as an example, if it cannot segment a region correctly in the vertical direction, the horizontal sub-model could complement the unsatisfactory segmentation in the horizontal direction. Also, the vertical sub-model is required to help the horizontal sub-model in the same way. Thus, the two sub-models could complement each other to achieve the self-improvement until convergence.
Additionally, to make the training process more effective and efficient, we propose two sampling strategies (\emph{i.e.}, the \emph{basic sampling strategy} and the \emph{covering re-sampling strategy}). The former samples the discriminative patches and balances the different classes (\emph{i.e.}, tumor or non-tumor) to allow the efficient model training with less patch redundancy, while the latter guarantees the complementary help between different sub-models. These two strategies facilitate self-improvement for these sub-models together.

 Since our proposed method involves the sampled crossbar patches from two directions, and the cascaded training process to iteratively train the vertical and horizontal sub-models, we name our method as Crossbar-Net in the following parts. Overall, the contributions of our work can be summarized in the following four folds:
\begin{itemize}
\item Our crossbar patches could capture both the local detail information and global contextual information. Also, these patches are easy to sample without introducing any additional parameters to train.
\item Our cascaded training process could help provide complementary information between different sub-models to enhance the final segmentation. In our training process, the sub-models can perform the self-improvement iteratively.
\item Our model is easy to implement and extend. Beyond kidney tumor segmentation in CT images, we have evaluated our method on cardiac segmentation in MR images and breast mass segmentation in X-ray images, showing promising results and good generalization ability.
\item Our method is fast to train and test, although it is trained in a cascaded manner. Taking kidney tumor as an example, the training time is about 1h and the testing time is about 1.5s on a regular GPU.
\end{itemize}

The rest of this paper is organized as follows. Related works about kidney tumor segmentation in recent years are briefly introduced in Section \ref{sec:related-work}. We then describe the framework and technical details of Crossbar-Net in Section \ref{sec:Method}. Experimental results are reported in Section \ref{sec:experiments}. Finally, we conclude our paper in Section \ref{sec:conclusion}.

%figure framework
\begin{figure*}[htb]
\setlength{\belowcaptionskip}{-2cm}
\centering
 \includegraphics[width = 6.8in]{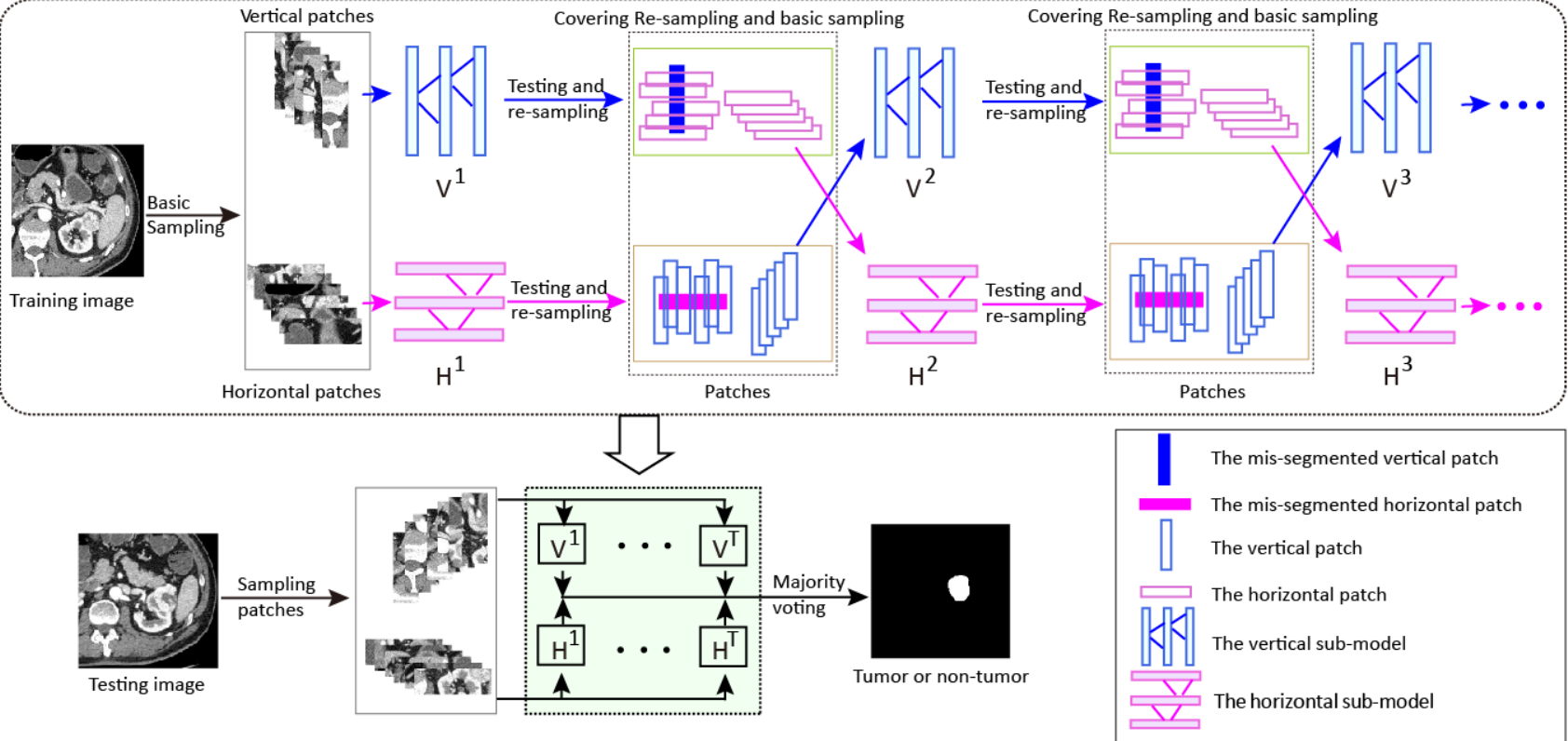}
\caption{Framework of the proposed method. $\mathbf{V}^i$ and $\mathbf{H}^i$ are vertical sub-model and horizontal sub-model of the $i$-th round, respectively, $i=1, \cdots, T$.}\label{framework}
%\vspace{-0.2cm}
\end{figure*}

\section{Related Work}
\label{sec:related-work}
For kidney tumor segmentation, according to the way of feature representation, most of the previous methods belong to the low-level methods. The low-level methods either employ the energy minimization-based models or learn the segmentation model on the extracted hand-crafted features. For example, Skalski \emph{et al.} \cite{skalski2016kidney} first located the kidney region through a hybrid level set method with the ellipsoidal shape constraint, then calculated the low-level features (\emph{e.g.}, mean value, standard deviation, histogram of oriented gradients \cite{tsai2010histogram}) on the obtained region, and finally performed the decision tree to distinguish the kidney tumor and arterial blood vessel regions. Linguraru \emph{et al.} \cite{linguraru2011automated} first extracted kidney tumors by the region growing for initial segmentation, and then applied geodesic active contours to refine the segmentation result. Also, Linguraru \emph{et al.} \cite{linguraru2009computer} described the kidney tumors using the low-level visual features ({\emph{e.g.}, histograms of curvature features). Similarly, Lee \emph{et al.} \cite{lee2017detection} first detected region-of-interest by analyzing the textural and contextual information, and then extracted the mass candidates with the region growing and active contours. Hodgdon \emph{et al.} \cite{hodgdon2015can} first extracted the texture features (\emph{i.e.}, gray-level histogram mean and variance, gray-level co-occurrence and run-length matrix features \cite{tang1998texture}), and then trained a support vector machine (SVM) to segment the fat-poor angiomyolipoma from renal cell carcinoma in CT images. These aforementioned low-level methods perform well in the simple case when the tumors show different appearance with surrounding tissues. However, their performance could not be fully guaranteed when the shape and texture of the tumors are close to the surrounding tissues.

Recent trends of using deep features or deep models have demonstrated the effectiveness in several segmentation tasks. Although the attempts of developing the specific deep feature-based methods for segmenting kidney tumor are very limited, the related deep feature-based segmentation methods for segmenting other medical organs \cite{ciresan2012deep},\cite{wang2017central, shi2017does, moeskops2016automatic, prasoon2013deep}, can be borrowed to segment the kidney tumor. For instance, Ciresan \emph{et al.} \cite{ciresan2012deep} employed multiple deep networks to segment biological neuron membranes by extracting the squared patches in multi-scales with sliding-window. Prasoon \emph{et al.} \cite{prasoon2013deep} designed a triplanar CNN model to segment cartilage with multiple view patches. Moeskops \emph{et al.} \cite{moeskops2016automatic} proposed a multi-scale CNN model to segment brain MR images. Wang \emph{et al.} \cite{wang2017central} devised a multi-branch CNN model to segment lung nodules. Shi \emph{et al.} \cite{shi2017does} proposed a cascaded deep domain adaptation model to segment the prostate in CT images. However, we notice that the squared patches used in these above works are one of the major bottlenecks when borrowing them to segment the kidney tumors which are experimentally demonstrated in our experiment.
To address the limitation of the local squared patch, several attempts about the image-level segmentation are exploited in recent years. Mortazi \emph{et al.} \cite{mortazi2017cardiacnet:} presented a novel multi-view CNN model to fuse the information from the axial, sagittal, and coronal views of cardiac MRI. This model performed well in the task of the left atrium and proximal pulmonary veins segmentation. Ronneberger \emph{et al.} \cite{ronneberger2015u} proposed a widely-used model in medical image segmentation tasks, namely U-Net, to solve the cell tracking problem. Recently, a new image-based model, SegCaps \cite{lalonde2018capsules} was developed which achieved a promising result in the task of segmenting pathological lungs from low dose CT scans. Also, He \emph{et al.} \cite{he2018pelvic} introduced a fully convolutional network with distinctive curve to segment the pelvic organ.

In summary, compared with the previous segmentation methods, all of them employ either the image-level or squared patch-level segmentation, while our Crossbar-Net involves the crossbar patches (non-squared patches) to capture the both (1) local detail information and (2) global context information from vertical and horizontal directions.

Recently, training deep models in a cascaded or boosting-like manner for better performance have aroused considerable interests \cite{Elad2016learning, Karianakis2015boosting, Schwenk2000boosting, Shai2017SelfieBoost, havaei2017brain}. For example, Shwartz \emph{et al.}\cite{Shai2017SelfieBoost} proposed a SelfieBoost model, which boosted the performance of a single network based on minimizing the maximal loss. Karianakis \emph {et al.} \cite{Karianakis2015boosting} proposed an object detection framework according to the boosted hierarchical features. Walach \emph{et al.} \cite{Elad2016learning} introduced a boosted-CNN model where the latter CNN was added according to the error of the former CNN, and finally all CNNs were joined via a sum layer. Similarly, Havaei \emph{et al.} \cite{havaei2017brain} presented a cascaded architecture consisting of two CNNs to segment glioblastomas in MR images, where the output probabilities of the first CNN was added to the layers of the second CNN. For these above methods, multi-modal fusion and enhancement in a boosting-like manner are the common choices. Also, our Crossbar-Net consists of the fusion from vertical and horizontal sub-models. As for what to enhance, in addition to hierarchical features \cite{Karianakis2015boosting, havaei2017brain}, the misclassified samples are enhanced in the next rounds to raise the concerns from the classifier. Adaboost \cite{freund1999short} and co-training models \cite{blum1998combining, Goldman00enhancingsupervised, muslea2000selective, zhou2006enhancing, Ehsan2017cotraining} are the typical misclassified-samples-enhancing methods. Inspired by this, we train a cascaded model composed of vertical and horizontal sub-models by enhancing the region around misclassified pixels.

Compared with the previous cascaded methods, the major distinctions of Crossbar-Net are (1) our model learns both local detail features and global context features from two directions simultaneously, and (2) our model is composed of the sub-models from two directions, in which the sub-models can perform the self-improvement during different rounds to complement each other iteratively.

\section{Method}
\label{sec:Method}
In this section, we first introduce our methodology and sampling strategy of crossbar patch, then present the sub-model setup and illustrate the training process, and finally discuss the testing process.

\subsection{Our Methodology}
\label{subsec:crossbar-net}
The framework of Crossbar-Net is schematized in Fig. \ref{framework}, which includes the training and testing stages.
Note that, we convert our segmentation task to a pixel-wise classification problem, which intends to predict a patch to be a tumor or non-tumor class.

In the training stage, firstly, the crossbar patches are initially extracted from the training CT images under the \emph{basic sampling strategy} (detailed in Section \ref{subsec:strategy}) with the manual segmentation available as the ground truth. Also, the vertical and horizontal sub-models are initially trained in the 1-st round, denoted as $\mathbf{V}^1$ and $\mathbf{H}^1$, respectively. Then, regarding the cascaded training process, at the $t$-th round, we evaluate the segmentation performance of the current trained vertical and horizontal sub-models, and select the mis-segmented regions of each sub-model. Formally, the mis-segmented region indicates the vertical or horizontal patch whose central pixel is misclassified, \emph{i.e.}, if a central pixel along with its located vertical patch is misclassified by a vertical sub-model, then its corresponding vertical patch is a mis-segmented region. And then, we re-sample the mis-segmented regions using the \emph{covering re-sampling strategy} (detailed in Section \ref{subsec:strategy}) to obtain the corresponding re-sampling patches. Then, we feed these re-sampling patches to another sub-model for its model training. Meanwhile, beyond the aforementioned re-sampling patches, in each round, the patches sampled under the \emph{basic sampling strategy} are also feed to the same sub-model. We keep repeating the above process until the training error converges or the maximum round number reaches.

In the testing stage for segmenting a new coming CT image, the trained sub-models in each round are gathered together to perform a majority voting on this image to obtain the final segmentation.
%figure basic sampling
\begin{figure}[htb]
\setlength{\belowcaptionskip}{-2cm}
\centering
\subfigure[]{
 \label{radius_example:a}
 \includegraphics[width=1.25in]{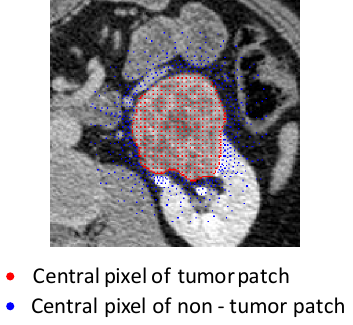}
}
\subfigure[]{
 \label{radius_example:b}
 \includegraphics[width=1.9in]{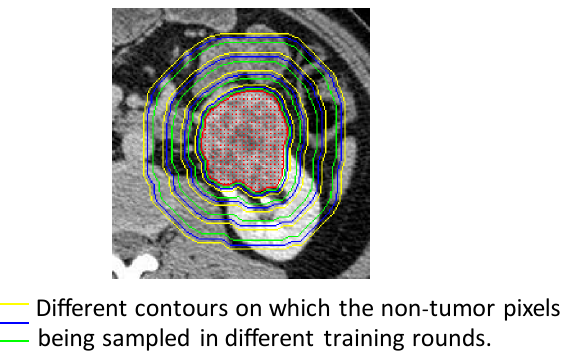}
}
\caption{The typical example of \emph{basic sampling strategy}.}\label{radius_example}
\vspace{-0.2cm}
\end{figure} %re-sampling strategy

\subsection{Crossbar Patch Sampling Strategy}
\label{subsec:strategy}
\subsubsection{Basic Sampling Strategy}
\label{subsec:basic}

We develop the \emph{basic sampling strategy} with the goal of making the segmentation model more focus on the region surrounding the tumor boundary which is considered to be hard to segment in practice \cite{shi2015semi}. Therefore, the principle we adhere to is to increase the patches close to the tumor and reduce the redundant patches that are far from the tumor. This sampling strategy is used in each round of the cascaded training with different sampling intervals.

By using this strategy, we select a part of total pixels according to the distance between the current pixel and the center of the tumor. We first extract crossbar patches uniformly in the tumor region as the training samples belonging to the tumor class (\emph{i.e.}, tumor patch), and then sample non-tumor patches densely near the tumor and sparsely in the far region as the training samples belonging to the non-tumor class (\emph{i.e.}, non-tumor patch). As shown in Fig. \ref{radius_example:a}, the red pixels are the center of the tumor patches with certain intervals and the blue pixels are the center of the non-tumor patches. More details are described in Section \ref{subsec:cascading}.
%figure
\begin{figure}[tb]
\centering
\setlength{\belowcaptionskip}{-1cm}
\includegraphics[width =3.4in]{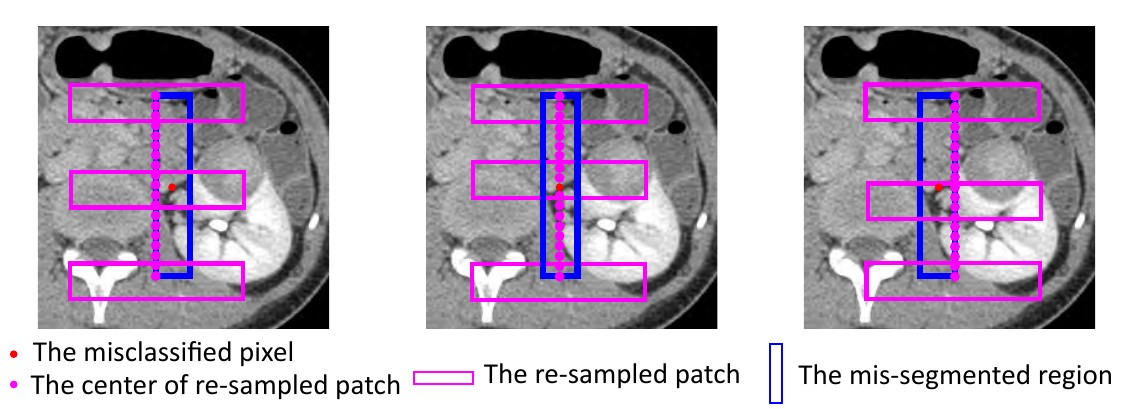}
\caption{Illustration of \emph{covering re-sampling strategy}. The blue patch is the mis-segmented region with the red central pixel, and the magenta rectangles are the re-sampling patches with central pixels being located at the left, middle and right line of the blue patch.}\label{re-sampling}
\vspace{-0.2cm}
\end{figure}

\subsubsection{Covering Re-sampling Strategy}
\label{subsec:resample}
As shown in Fig. \ref{re-sampling}, we assume that the vertical patch is a mis-segmented region in vertical sub-model $\mathbf{V}^t$ at the $t$-th round. Our purpose is to borrow the horizontal sub-model to well segment this region using the horizontal patches. In particular, we extract the horizontal patches by fully covering the mis-segmented region, according to the location of the misclassified central pixel in the vertical sub-model. In order to cover this region, we sample the horizontal patches with the central pixel being located at three columns: the center, right and left column of the vertical patch. To avoid sampling redundant horizontal patches, we perform re-sampling by every three pixels on each column. Normally, for a vertical patch, we can roughly obtain $\sim $ 40 horizontal patches. Thus, the horizontal sub-models can provide a complement to the vertical sub-models. Moreover, in this way, if both sub-models, \emph{i.e.}, $\mathbf{V}^t$ and $\mathbf{H}^t$, fail on a same pixel, the role of the region around this pixel will definitely be enhanced in the next round. Thus, with the combination of the re-sampling patches and the aforementioned basic-sampling patches as the training samples, the performance of $\mathbf{V}^{t+1}$ is expected to be superior to $\mathbf{V}^t$, and $\mathbf{H}^{t+1}$ is also expected to outperform $\mathbf{H}^t$.
%table 1
\begin{table*}[tb]
\begin{center}
\caption{Details of sub-model architectures. Conv and Pooling denote convolutional layer and pooling layer, respectively.}
\label{arch_details}
{\scriptsize
\renewcommand{\arraystretch}{1}
\begin{tabular}{c|p{0.59in}|p{0.39in}p{0.36in}p{0.36in}p{0.26in}p{0.26in}p{0.26in}p{0.26in}p{0.26in}p{0.26in}p{0.26in}p{0.30in}}
\toprule
 & Layer &\# 1&\# 2&\# 3&\# 4&\# 5&\# 6&\# 7&\# 8&\# 9&\# 10&\# 11\\
\midrule
\multirow{3}{*}{Vertical sub-model} & Layer type & Conv & Conv & Pooling & Conv & Pooling & Conv & Conv & Conv & Conv & Conv & Softmax\\
& Feature maps & 16 & 36 & 36 & 64 & 64 & 64 & 64 & 64 & 500 & 2 & 2 \\
~& Kernel size &$5\times3$&$5\times3$&$2\times2$&$5\times3$ & $2 \times 2$ & $5\times 3$ & $6 \times 1$ & $6 \times 1$ & $7 \times 1$ & $1 \times 1$ & - \\
\midrule
\multirow{3}{*}{Horizontal sub-model} & Layer type & Conv & Conv & Pooling & Conv & Pooling & Conv & Conv & Conv & Conv & Conv & Softmax  \\
& Feature maps & 16 & 36 & 36 & 64 & 64 & 64 & 64 & 64 & 500 & 2 & 2 \\
~&  Kernel size & $3 \times 5$ & $3 \times 5$ & $2 \times 2$ & $3\times 5$ & $2 \times 2$ & $3 \times 5$ & $1 \times 6$ & $1 \times 6$ & $1 \times 7$ & $1 \times 1$ & - \\
\bottomrule
\end{tabular}}
\end{center}
\vspace{-0.2cm}
\end{table*}

\subsection{Sub-Model Architecture}
\label{subsec:architecture}
The architecture of sub-models is designed according to a preliminary study on our CT kidney tumor dataset. The number of layers, kernel size and the amount of feature maps are all experimentally determined by inner cross-validation. Basically, both vertical and horizontal sub-models consist of eight convolutional layers, two max-pooling layers, and one softmax layer. Details of sub-model are illustrated in Table \ref{arch_details}. For the vertical sub-model, the input is a $100 \times 20$ vertical patch. Regarding the patch is non-square, the sizes of convolutional kernels in the first four convolutional layers are all set to $5 \times 3$, while that in the last four convolutional layers are set to $6 \times 1$, $6 \times 1$, $7 \times 1$ and $1 \times 1$, respectively. Each convolutional layer is followed by the rectified linear unit (ReLU) \cite{nair2010rectified} activation and performed with 1 stride and 0 padding. The kernel size of each pooling layer is $2 \times 2$ with 2 strides and 0 padding. In addition, the dropout \cite{hinton2012improving} after the last convolutional layer is applied to avoid the possible over-fitting. For each sub-model, we minimize the following softmax loss $L$:
\begin{equation}
L=-\sum_{i=1}^{N}y_{i}\text{log}\widehat{y}_{i} \label{con:loss}
\end{equation}
where $N$ is the number of training patches (vertical or horizontal patches) in the current sub-model, $y_{i}$ and $\widehat{y_{i}}$ are the ground truth and the predicated label of the central pixel in the $i$-th patch, respectively. The weights of the filters are initialized randomly with the Gaussian distribution \cite{lecun2012efficient} and updated by the stochastic gradient descent (SGD) algorithm.

Similarly, we can also obtain the architecture of the horizontal sub-model as illustrated in Table \ref{arch_details}, with the input of the sub-model as a $20 \times 100$ horizontal patch. Please note that the architecture could be adjusted according to different segmentation scenarios. Typically, we use the same architecture in the cardiac segmentation and different architecture in the breast mass segmentation.
\subsection{Training Sub-Models}
\label{subsec:cascading}

We now discuss how to train our Crossbar-Net in a cascaded manner, as a boosting-like training style. Formally, we denote the vertical and horizontal sub-models in the $i$-th round as $\mathbf{V}^i$ and $\mathbf{H}^i$, respectively. The training process can be detailed as follows:

\textbf{Firstly}, extracting crossbar patches with the \emph{basic sampling strategy} and training the initial vertical and horizontal sub-models, \emph{i.e.}, $\mathbf{V}^1$ and $\mathbf{H}^1$, respectively. We use an example to illustrate the detailed implementation process of \emph{basic sampling strategy}:

As shown in Fig. \ref{radius_example:b}, in this round, we first sample pixels on the green contours as the centers of non-tumor patches. Then, we select the red points as the centers of tumor patches on the odd rows inside the tumor. The sampling interval on the contours near the tumor is smaller than that on the contours which are far away from the tumor, with the goal of sampling more patches near the tumor boundary. Here, we denote the sets of these pixels (\emph{i.e.}, the location of these pixels) and their ground truth labels as $X$ and $Y$, and the corresponding vertical and horizontal training patches as $P_{\text{basic}}^{V^{1}}$ and $P_{\text{basic}}^{H^{1}}$, respectively.

\textbf{Secondly}, we continuously update our Crossbar-Net based on the currently obtained sub-models. Specifically, in the $i$-th round ($i>1$),
\begin{itemize}
\item Performing the evaluation on $\mathbf{H}^{i-1}$. In particular, we input the $P_{\text{basic}}^{H^{1}}$ to $\mathbf{H}^{i-1}$ for the evaluation at the $i$-th round, by predicting the label of central pixel in each patch from $P_{\text{basic}}^{H^{1}}$. In this meaning, the mis-segmented regions in $\mathbf{H}^{i-1}$ are determined according to the predicated labels.
    Formally, assuming that all these predicted labels as $\widehat{Y}_{H}$, we define the misclassified pixels as:
    \begin{equation}
    C_{H}^{i-1}=\Bigg\{ x_{j}|x_{j}\in X \wedge I(\widehat{y}_{j}\neq y_{j}), j=1,...,N \Bigg\} \label{con:mis}
    \end{equation}
    where $x_{j}$ is the central pixel of the $j$-th patch, $\widehat{y}_{j}$ and $y_{j}$ are the predicted and ground truth label of $x_{j}$, $\widehat{y}_{j}\in \widehat{Y}_{H}$ and $y_{j}\in Y$. $N$ is the number of horizontal training patches. $I(s)$ is an indicator function. $I(s)=1$ if and only if the statement $s$ is true and $I(s)=0$ otherwise.

\item Performing both \emph{covering re-sampling strategy} (on the mis-segmented regions in $\mathbf{H}^{i-1}$) and the \emph{basic sampling strategy}, to obtain the newly generated vertical patches.

    Firstly, the covering re-sampling is performed. We also record the position of the patches obtained by \emph{covering re-sampling strategy} in the current round, denoted as $P_{\text{re}}^{V^{i}}$. Then, the \emph{basic sampling strategy} is sequentially performed. Specifically, as $i$ varies, as shown in Fig. \ref{radius_example:b}, we sample central pixels for non-tumor patches on different contours which are labeled by different colors. For the tumor patches, we sample the red points on different rows or columns as the central pixels. The sampling intervals for these two types of patches are different from that in the previous rounds. Meanwhile, if a patch has already been extracted in the covering re-sampling, it cannot be sampled in the basic sampling in the same round. The advantage of the above way is to avoid redundant sampling which might cause over-fitting.
    Here, the \emph{covering re-sampling strategy} aims to enhance the role of mis-segmented regions, while the \emph{basic sampling strategy} wishes to control the amount and distribution of training samples and prevent the sub-model from over-emphasizing the misclassified pixels.
\item Employing these newly generated patches to train $\mathbf{V}^{i-1}$ to obtain $\mathbf{V}^{i}$.
\item Updating the $\mathbf{H}^{i}$ from $\mathbf{H}^{i-1}$, similarly.
\end{itemize}

\textbf{Finally}, we repeat the aforementioned steps by updating $\mathbf{H}^{i}$ and $\mathbf{V}^{i}$ from $\mathbf{H}^{i-1}$ and $\mathbf{V}^{i-1}$ iteratively, until (1) maximum training round number reaches or (2) the training error of each sub-model could not be reduced significantly anymore. 	

Overall, the advantages of our cascaded training can be summarized as: The vertical and horizontal sub-models could complement each other during the cascaded training. When the features in one direction are not very discriminative for segmentation, the features in another direction could make up for the current direction. For example, if the boundaries are blurred in the vertical direction, they might be sharp in the horizontal direction. In the remaining rounds, the vertical (horizontal) sub-model iteratively feeds the generated crossbar patches using \emph{covering re-sampling strategy} to the horizontal (vertical) sub-model in the same round, until the convergence. This can emphasize the learning on the mis-segmented regions and guarantee the sub-models to complement each other.

As a boosting-like algorithm, both the horizontal and vertical sub-model can perform self-improvement with both \emph{basic sampling patches} and \emph{covering re-sampling patches} as the training samples. Here, the self-improvement means that the performance of sub-model in one direction can be improved along with the increase of rounds. We claimed that the \emph{basic sampling patches} in the current round are sampled at different intervals with the previous rounds, which is equivalent to adding new training data to the sub-model. This might be a major cause of the self-improvement. In addition, if both sub-models in the same round fail on a same pixel, the corresponding mis-segmented region of this pixel in the current round will be definitely enhanced in both vertical and horizontal sub-models in the next round. In this meaning, the region around this misclassified pixel could have a larger chance to be emphasized in the next round compared with the current round, which causes the segmentation model in the next round more cares about the segmentation error on this region. Thus, a better segmentation performance on this region is expected as the similar weight-updating way in AdaBoost.

\subsection{Testing }
\label{subsec:testing}
In the testing phase, for a new coming image, we first extract the crossbar patches for each pixel. Then, we input these extracted patches to the trained vertical and horizontal sub-models in each round, to predict the central pixel of each patch belongs to the tumor region or not. Each sub-model outputs a segmentation result, and the final result is generated by a majority voting on all obtained results. Formally, assuming $T$ as the number of maximum round, we can obtain $2T$ sub-models as $\mathbf{V}^1, \cdots, \mathbf{V}^{T}$ and  $\mathbf{H}^1, \cdots, \mathbf{H}^{T}$. The result is determined by these $2T$ sub-models.

\section{Experimental Results}
\label{sec:experiments}
Now we validate the advantages of Crossbar-Net qualitatively and quantitatively. After the introduction of datasets, evaluation criteria, and implementation details, we first investigate the characteristics of Crossbar-Net. Then, we present comparisons between Crossbar-Net and baseline methods on kidney tumor dataset. Finally, we apply Crossbar-Net to the cardiac and breast mass segmentation task to show that our model could be extended to other organ segmentation.

\subsection{Data}
\textbf{Kidney tumor dataset}. This dataset is independently collected from Suzhou Science and Technology Town Hospital. 3,500 CT slices of 94 subjects in total are used for performance evaluation, with one tumor per slice. The resolution of the image is $512 \times 512$ with  $1 \times 1$ mm$^2$/pixel, and the spacing between slices is 1 mm. For each image, the diameter of tumors ranges from 7 pixels to 90 pixels, and the tumor is manually annotated by the physician as the ground truth for training. We randomly partitioned the dataset into three subsets including the training, validation and testing sets which consist of 50, 8 and 36 subjects respectively. The sub-models in the 1-st round is trained using about 580,000 patches being extracted by the \emph{basic sampling strategy}. The epoch number of each sub-model is automatically determined by the validation set, and the performance of each sub-model in each round is still evaluated by the training set.

\textbf{Breast masses dataset}.
A subset of DDSM \cite{Heath2000digital, Heath1998current} is introduced to investigate the performance of our method. The image in this dataset is LJPEG format and we convert them into PNG format as \cite{Anmol2015}. There are 1,923 malignant and benign cases in DDSM, and each case includes two images of each breast. The Regions of Interesting (ROI) are given in images containing the suspicious areas. Since the ROIs are not the accurate boundary of a tumor, the boundary of each tumor is annotated again as the ground truth by the experienced radiologists. There are in total 1,000 selected images, among which 600 and 100 images are training and validation set and the remaining 300 images are testing set. In most of the breast mass segmentation methods, the image is cropped to the bounding box of ROI \cite{dhungel2015deep, zhu2018Adversarial} or 1.2 times of the bounding box \cite{cardoso2017mass}. We follow the cropping manner in \cite{cardoso2017mass}. An example is shown in Fig. \ref{breastcropping:a}. According to the \emph{basic sampling strategy}, some patches might be extracted outside the image (Fig. \ref{breastcropping:b}). For these patches, the parts outside the image are filled with black. When implementing the \emph{covering re-sampling strategy}, the black part of mis-segmented regions would not be re-sampled. Most tumors in these images are smaller than 450 pixels in diameter with very few of them whose diameter is about 600 pixels.

\textbf{Cardiac dataset}.
This dataset is available from \cite{andreopoulos2008efficient} and comprised of cardiac MRI sequences for 33 subjects with total 7,980 MR slices. The image resolution is $256 \times 256$ with the in-plane pixel size as 0.9-1.6 mm$^2$, and the inter-slice thickness as 6-13 mm. In each image, endocardial and epicardial contours of the left ventricle (LV) are provided as the ground truth. We randomly select 20 and 3 subjects as the training and validation set to train sub-models. The images of the remaining 10 subjects form a testing set for evaluation. The amount of training patches is about 350,000 in the 1-st round, and about 100,000 and 50,000 samples are sampled in the 2-nd and 3-rd round respectively.

%figure breast mass cropping
\begin{figure}[htb]
\setlength{\belowcaptionskip}{-2cm}
\centering
\subfigure[]{
 \label{breastcropping:a}
 \includegraphics[width=1.6in]{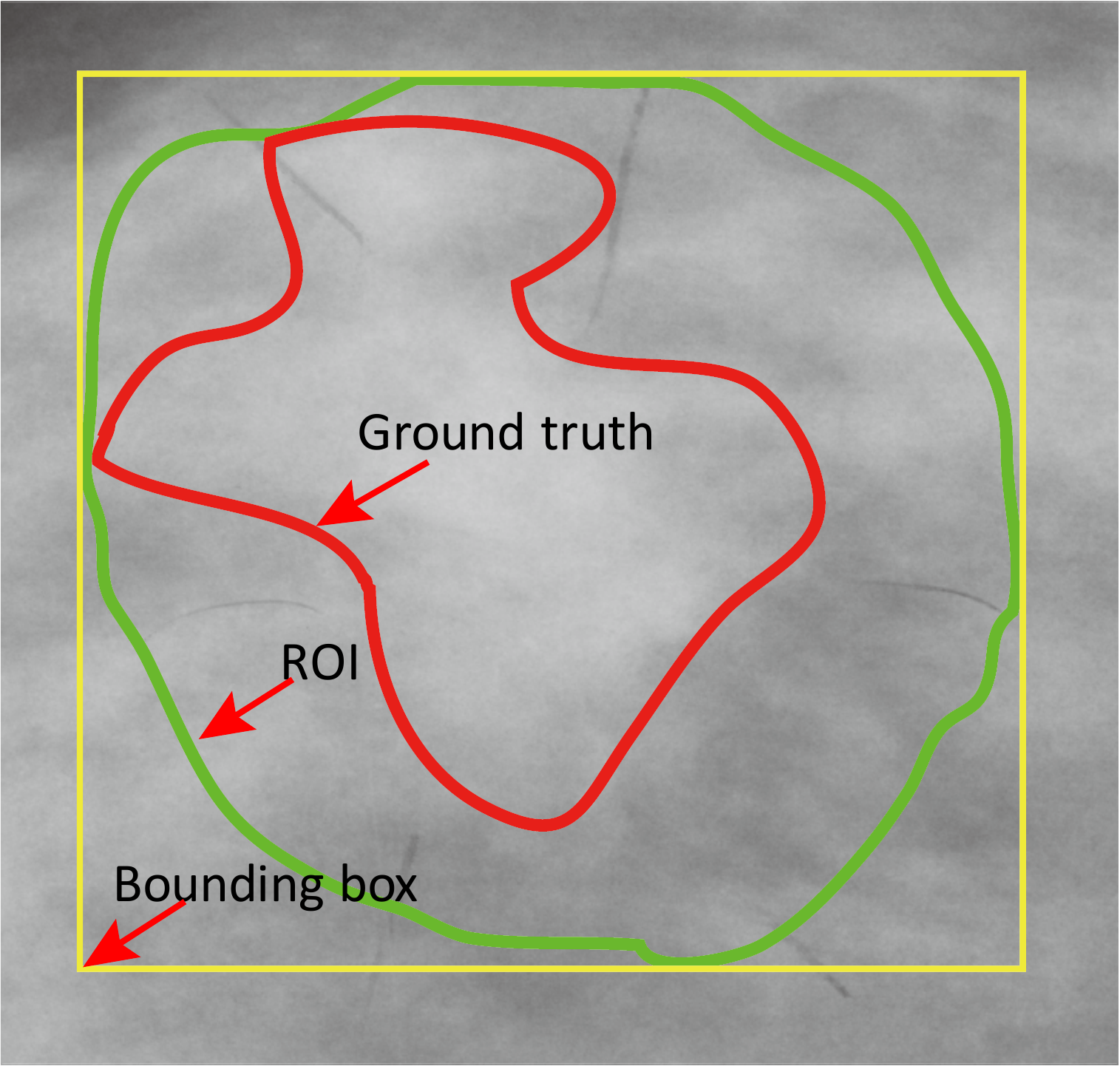}
}
\subfigure[]{
 \label{breastcropping:b}
 \includegraphics[width=1.6in]{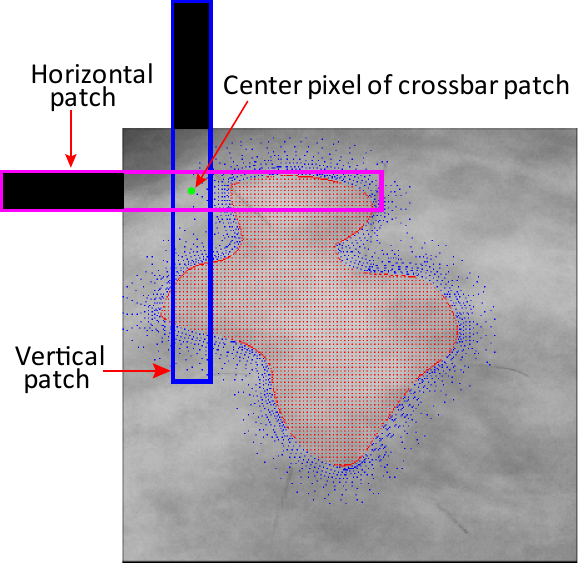}
}
\caption{Example of the cropped breast mass image and the extracted crossbar patch. (a) is the cropped image. (b) shows the crossbar patch extracted outside the image.}\label{fig:breastcropping}
\end{figure}

 \subsection{Evaluation Criteria}
We employ the Dice similarity coefficient (DSC) and the true positive fraction (TPF) as the primary evaluation criteria for assessing the segmentation performance. DSC is usually employed to measure the overlap between the prediction and manual segmentation. A large DSC indicates a high segmentation accuracy. TPF indicates the percentage of correctly predicted tumor pixels in the manually segmented region. The higher the TPF is, the larger the coverage of the true tumor region is. We also introduce the centroid distance (CD) and the Hausdorff distance (HD) to evaluate the segmentation accuracy. CD indicates the distance between the central pixels of the final segmentation and manual segmentation which is used to indicate the Euclidean distance between two central points in a 3-D space. Similar to CD, a smaller HD indicates higher proximity between the final segmentation and manual segmentation, which is introduced in quite a bit of detail in Zhang \emph{et al.} \cite{zhang2015deep}. More details about DSC, TPF, and CD are introduced in \cite{shi2016learning}.

\subsection{Implementation Details}
 For the scale of crossbar patch on all datasets, we set the size of the horizontal patch as $20 \times 100$ and vertical patch as $100 \times 20$ on kidney data and cardiac data, and $50 \times 500$ and $500 \times 50$ on DDSM, respectively. We implement our networks on MatConvnet toolbox \cite{vedaldi2015matconvnet}. In order to improve the credibility of segmentation, the training and testing procedure are repeated 3 times in all experiments. In each time, the training, validation, and testing subsets are selected randomly, and we report the final average performance. Each sub-model reaches convergence within 20, 25 and 25 epoches on kidney tumor, cardiac and DDSM, respectively. We run all deep models on NVIDIA GTX 1080 Ti.
%figure error
\begin{figure}[tb]
\centering
\includegraphics[width = 2.8in]{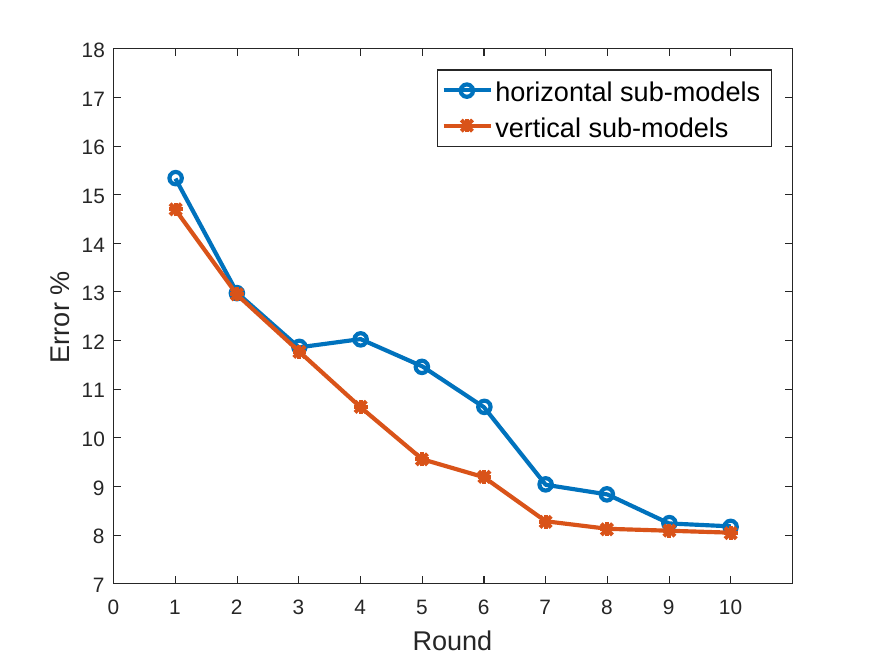}
\caption{Illustration of training error of each sub-model. The sub-models are trained separately.}\label{error-of}
\end{figure}

%figure stage submodel
\begin{figure}[tb]
\centering
\includegraphics[width = 3.4in]{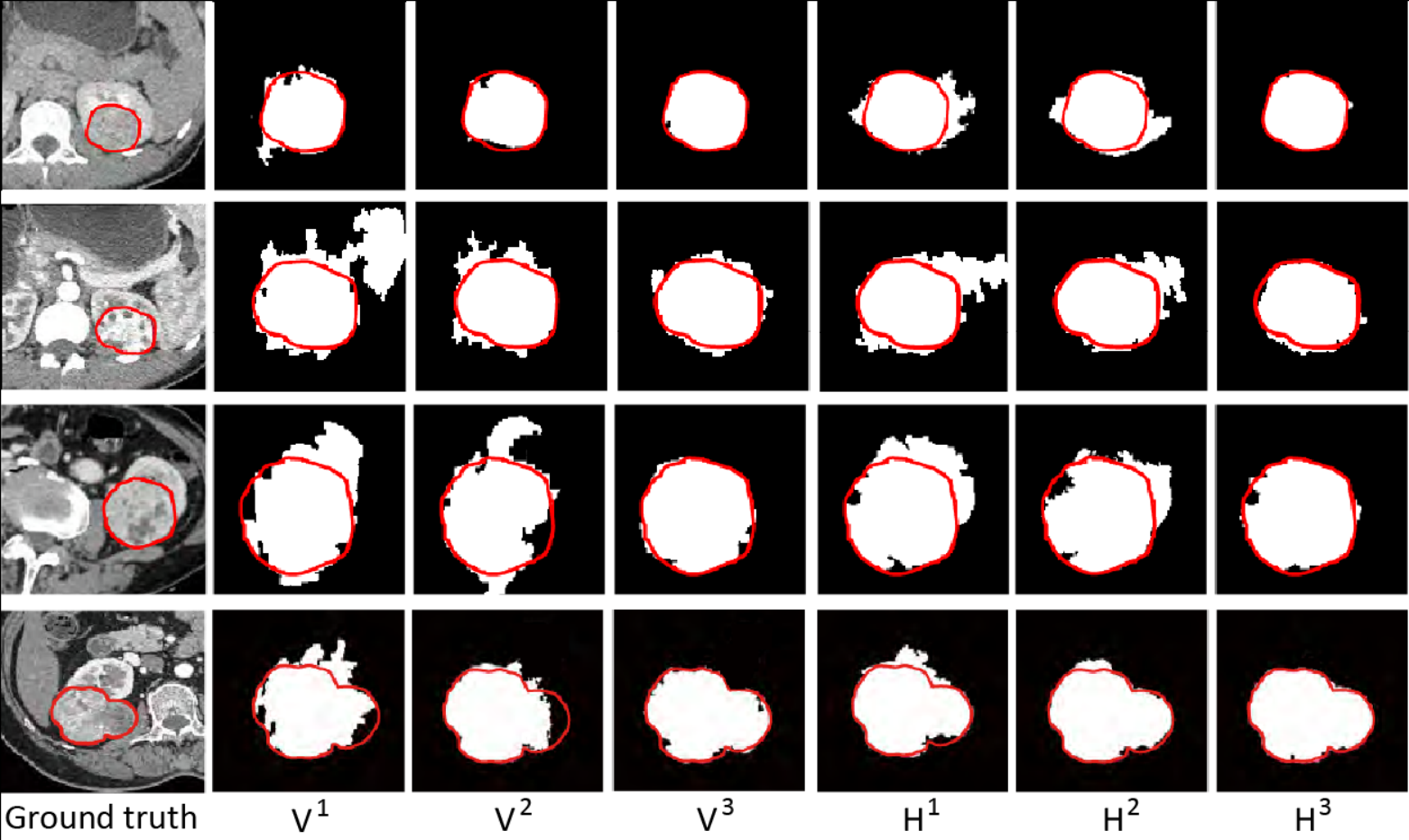}
\caption{Performance of each sub-model on kidney tumor. The left column indicates the ground truth image, the second to forth columns indicate the results of three vertical sub-models, and the last three columns indicate the results of three horizontal sub-models, respectively.}\label{stage_example}
\end{figure}
%%%%%%%%%%%%%%%%%%%%%%%%%%%%%%%%%%%%%%%%%%%%%%%%%%%%%%%%%%%%%%%%%%%%%%%
%figure DSC FPT
\begin{figure}[tb]
\centering
\subfigure[under different patches]{
 \label{DSC:a}
 \includegraphics[width=1.6in]{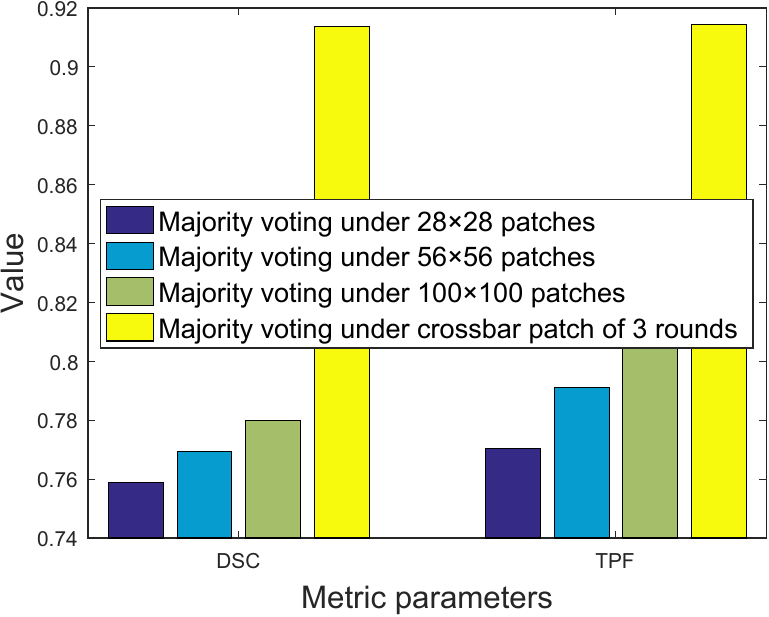}
}
\subfigure[under crossbar patches]{
 \label{DSC:b}
 \includegraphics[width=1.6in]{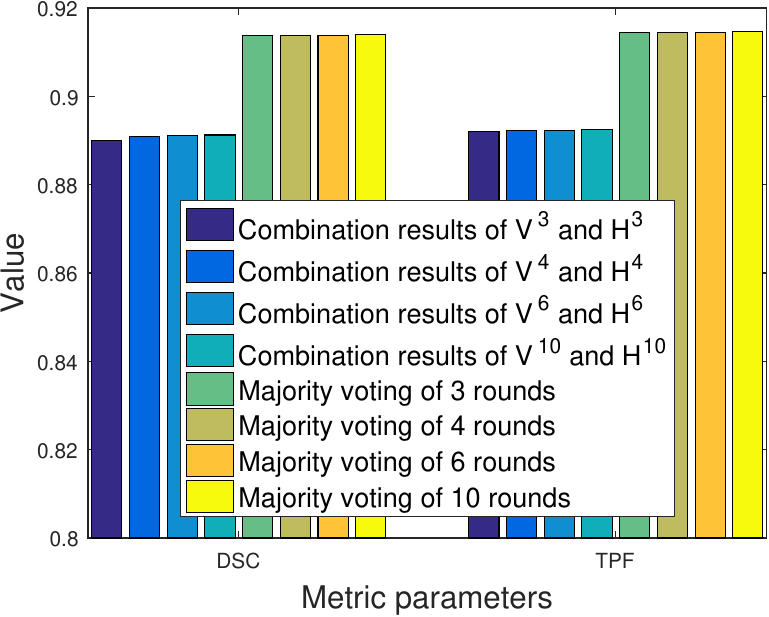}
}
\caption{DSC and TPF of Crossbar-Net.}\label{dr_and_tpf}
\end{figure}
%%%%%%%%%%%%%%%%%%%%%%%%%%%%%%%%%%%%%%%%%%%%%%%%%%%
 %table3
\begin{table}
\fontsize{7.5}{8}\selectfont
\setlength{\belowcaptionskip}{-0.2cm}
\caption{Average metrics for each sub-model on testing set}
\label{average_metrics}
\begin{center}
\renewcommand\arraystretch{1}
\begin{tabular}{p{0.35in}|p{0.29in}p{0.29in}p{0.29in}p{0.29in}p{0.29in}c}
\toprule
  &$\mathbf{V}^1$ &$\mathbf{V}^2$ &$\mathbf{V}^3$ &$\mathbf{H}^1$ &$\mathbf{H}^2$ &$\mathbf{H}^3$\\
\midrule
DSC&0.853&0.871&0.882&0.847&0.871&0.881\\
TPF&0.844 &0.876 &0.883&0.834&0.869&0.884\\
HD \tiny{(mm)} &10.200&9.586&9.223&11.315&10.001&9.890\\
CD \tiny{(mm)} &4.346&3.424 &2.790&4.403&3.560 &2.909\\
\bottomrule
\end{tabular}
\end{center}
\vspace{-0.2cm}
\end{table}
%%%%%%%%%%%%%%%%%%%%%%%%%%%%%%%%%%%%%%%%%%%%%%%%%%%%%%%%%%%%%%%%%%%%%%%%%%%%
%figure T-SNE
\begin{figure*}[tb]
\centering
\subfigure[The test image]{
 \label{tsne:a}
 \includegraphics[width=1.22in, height=1.22in]{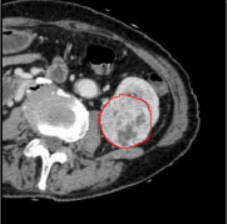}
}
\subfigure[Cluster of squared patches]{
 \label{tsne:d}
 \includegraphics[width=1.65in]{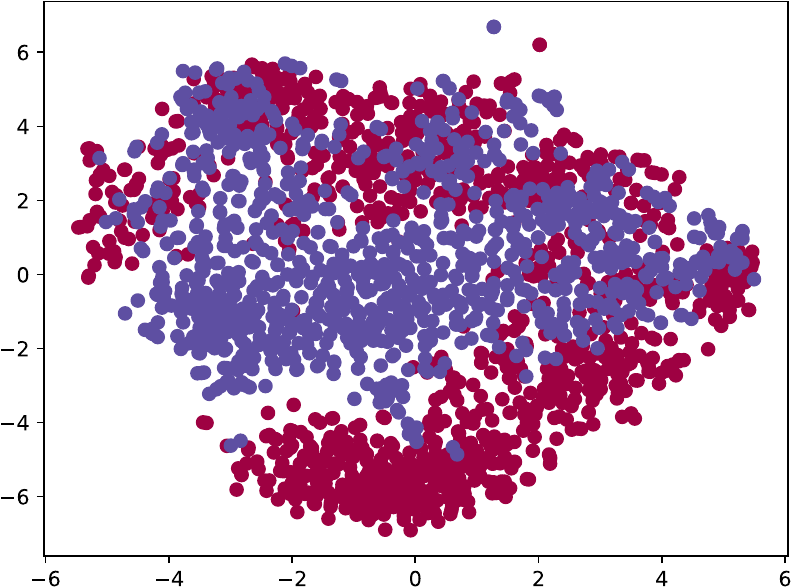}
}
\subfigure[Cluster of vertical patches]{
 \label{tsne:b}
 \includegraphics[width=1.65in]{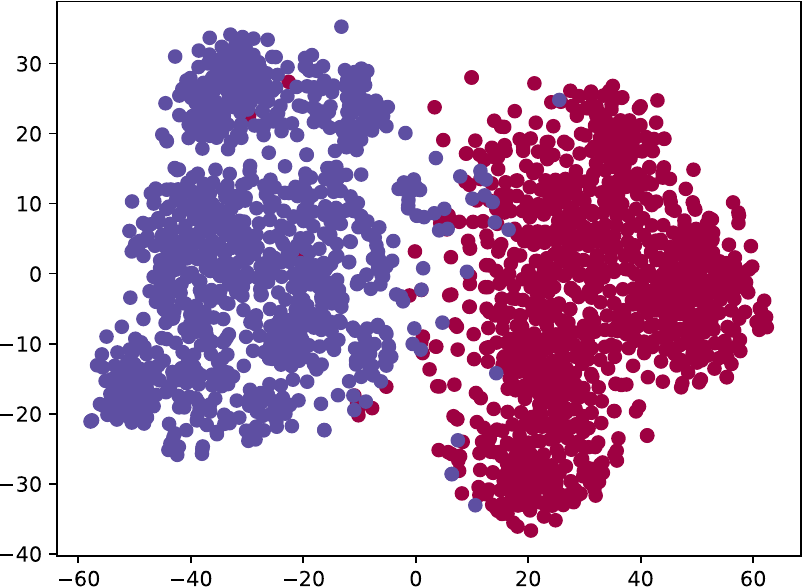}
}
\subfigure[Cluster of horizontal patches]{
 \label{tsne:c}
 \includegraphics[width=1.65in]{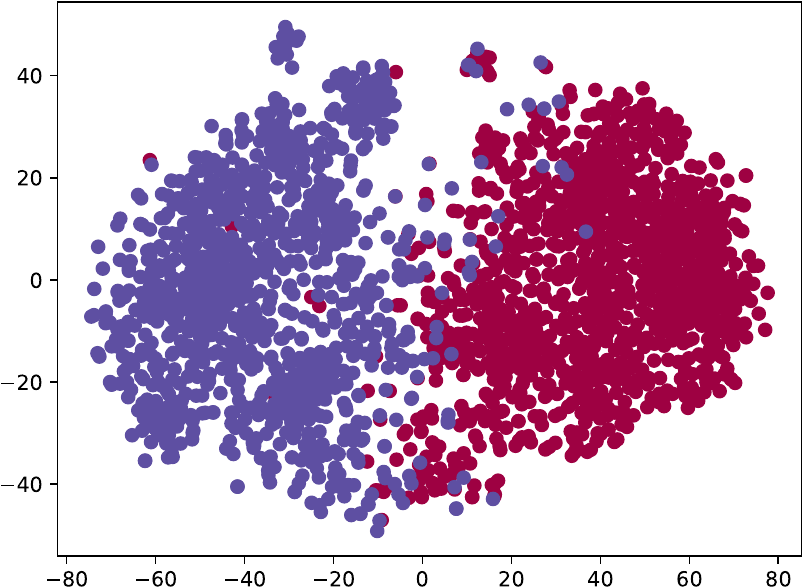}
}
\caption{t-SNE visualization of the high level representations in Crossbar-Net.}\label{tsne}
\end{figure*}

\subsection{Characteristics of Crossbar-Net}
We analyze four aspects about the characteristics of the proposed Crossbar-Net, which including: (1) the sub-model can perform self-improvement, (2) the sub-models in the same training iteration can complement and benefit each other, (3) the advantage of combining all the sub-models for final segmentation, and (4) the effectiveness of crossbar patches.

\textbf{Self-improvement}.
The purpose of this experiment is to show that if the vertical sub-model can self-improve without the involvement of the horizontal sub-model, and vice versa. We first train the vertical sub-model separately to obtain $\mathbf{V}^1$. Then, we re-sample the mis-segmented regions with the \emph{covering re-sampling strategy}, where the re-sampling patches are the vertical patches instead of the horizontal patches. Then, we train $\mathbf{V}^1$ with these patches together with those gotten from the \emph{basic sampling strategy} to get $\mathbf{V}^2$. Similarly, we can obtain the following $\mathbf{V}^3$, $\mathbf{V}^4$, $\cdots$ in a same way. We repeat training the sub-model (10 times here) until the training segmentation error converges. The horizontal sub-model is excluded throughout the process. The same process is also employed on horizontal sub-model. As illustrated in Fig. \ref{error-of}, the training error of each sub-model decreases with the increase of rounds. This experiment also shows that although each sub-model can self-improve, it takes 10 rounds for vertical (horizontal) sub-model to its convergence. In fact, only 3 rounds are needed if we train the sub-models in our manner shown in Fig. \ref{framework} instead of using this separate manner.

%%%%%%%%%%%%%%%%%%%%%%%%%%%%%%%%%%%%%%%%%%%%%%%%%%%%%%%%%%%%%%%%%%%%%%%
%figureKDE
\begin{figure}[tb]
\vspace{-0.3cm}
\centering
\includegraphics[width = 2.3in]{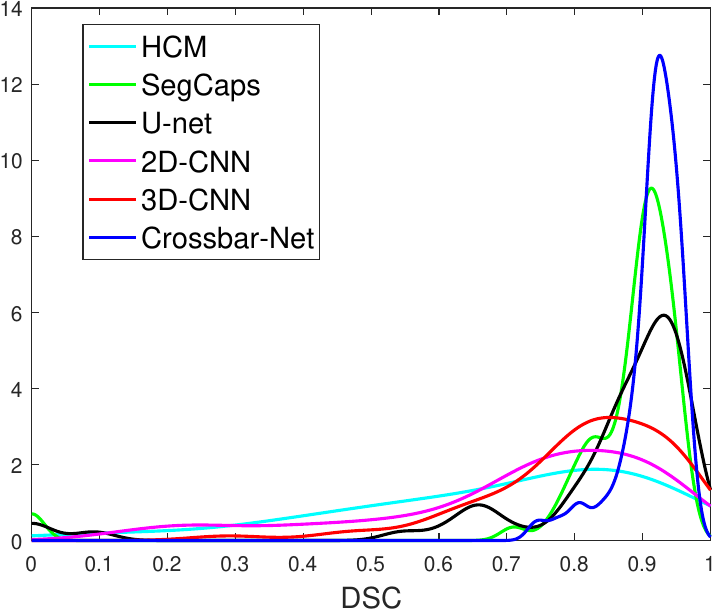}
\caption{Distribution of DSC on 600 kidney tumors.}\label{dr_bar}
\end{figure}

\textbf{Complement between Sub-models}. In this experiment, we train sub-models in the manner as illustrated in Fig. \ref{framework}. After 3 rounds, the training error of the vertical and horizontal sub-models reaches its convergence. Thus, the sub-models in these 3 rounds, which are $\mathbf{V}^1$, $\mathbf{V}^2$, $\mathbf{V}^3$ and $\mathbf{H}^1$, $\mathbf{H}^2$, $\mathbf{H}^3$, will be used in all the following kidney tumor experiments. As shown in Fig. \ref{stage_example}, we illustrate a typical case for visualization in the first row. In this case, we can observe $\mathbf{V}^1$ fails to properly segment the upper and lower parts of the tumor from vertical direction while $\mathbf{H}^1$ performs well from the horizontal direction. Similarly, $\mathbf{H}^1$ cannot segment the tumor with its left and right background correctly while $\mathbf{V}^1$ is better. At the same time, the degree of disagreement is reduced between $\mathbf{V}^2$ and $\mathbf{H}^2$. $\mathbf{V}^3$ and $\mathbf{H}^3$ both achieve the promising results eventually. For other tumors in the remaining rows, in addition to complementary regions, there are more or less common misclassified pixels. For example, in the second tumor, the upper right boundary and the lower right boundary are incorrectly segmented by both sub-models in the first two rounds. In this case, $\mathbf{V}^2$ and $\mathbf{H}^2$ are obviously superior to their former sub-models. Also, $\mathbf{H}^3$ and $\mathbf{V}^3$ achieves the best performance. This observation verifies that sub-models can benefit and complement each other. Basically, in all cases, the performances of later sub-models are superior to the former sub-models.

\textbf{The Way of Combination}. Table \ref{average_metrics} is the corresponding quantitative result of Fig. \ref{stage_example}, which confirms that the latter sub-model works better than the previous sub-model. However, we cannot simply combine the results of the two last sub-models. Intuitively, the latter sub model pays more attention to the difficult regions (\emph{i.e.}, hard to be segmented by the former sub-models). If we directly use the last sub-models as the segmentation model, we might have a strong attention-bias on the original data. Therefore, the majority voting of all sub-models is adopted, in which the weights of the last sub-models are greater than that of others. Experimentally, we indeed find the effectiveness of the majority voting. As shown in Fig. \ref{DSC:b}, the DSC and TPF of $\mathbf{V}^3$ and $\mathbf{H}^3$ combination are lower than the majority voting result of all sub-models in the 3 rounds around 2\%. Theoretically, our \emph{covering re-sampling strategy} is to some extent to being similar to the sampling strategy in AdaBoost. Also, in AdaBoost, the final strong ensemble classifier is obtained by a combination of all the previous weak classifiers since the latter weak classifier also assigns the higher weight to the difficult samples. We empirically set the weights of the last two models ($\mathbf{V}^3$ and $\mathbf{H}^3$) to 1.5 and the remaining to 1.

Although the training error of both vertical and horizontal sub-models converges after 3 rounds, to further show the benefit of combining all the sub-models with the majority voting, we continue the training process until reaching 10 rounds. The DSC and TPF of the 4, 6 and 10 rounds are listed in Fig. \ref{DSC:b} together with that of the 3 rounds. Obviously, as the number of training rounds increasing, the majority voting still remains superior to the simple combination of the last two sub-models.

\textbf{Effectiveness of Crossbar Patches}. We illustrate the advantages of crossbar patch by additionally comparing the crossbar patch with the squared patch (\emph{i.e.}, $28 \times 28$ and $56 \times 56$). For the experiment setting about these two types of patches, both their training strategy and network structure are maintained to be consistent for a fair comparison. For the training strategy, both of them adopt the cascaded training to make the network more focus on the mis-segmented regions, and their final segmentation models are the corresponding combination of their respective sub-models. For the network structure, it is impossible to make their networks be identical due to the different input sizes. However, for fair comparison, we tried our best to make their network structures to be almost same with only the difference on $1-3$ convolutional layers. Specifically, the structure in each path of the $28 \times 28$ and $56 \times 56$ patch is CCCPCCCCCS and CCPCCPCCCCCS respectively. Here C denotes convolutional layer, P is pooling layer and S is softmax layer. Since the filter is set to be rectangular for the rectangular patch, it is natural to set the filter to be square if the patch is square. Therefore, for the kernel size in squared patches, the $3 \times 3$ filter is adopted in all convolutional layers of all sub-models. The training data of both vertical and horizontal sub-models are sampled on the same images, while sampling intervals are different.

The results are shown in Fig. \ref{DSC:a}. The DSC and TPF of the model using squared patches are much lower than that using crossbar patches. In order to verify the unappealing result is not caused by the small size of the input patch, we turn the size of patches into $100 \times 100$. The structure of sub-model in each path under $100 \times 100$ patches is CCPCCPCCPCCCCS. The results are shown in Fig. \ref{DSC:a}, without any significant improvement. This observation indicates that, compared with non-squared crossbar patch, the large squared patches may include more information, while they might also bring in irrelevant noise to distinguish boundary.

Furthermore, in order to highlight the effectiveness of the non-squared patch, we also compare the features learned from crossbar patch and squared patch. We use t-SNE (t-distributed Stochastic Neighbour Embedding) \cite{maaten2008visualizing} to evaluate the features. We take the 500-dimensional features in $\mathbf{V}^3$ from the vertical patch and the $56 \times 56$ patch, in $\mathbf{H}^3$ from the horizontal patch, respectively. As shown in Fig. \ref{tsne}, each point represents a patch projected from 500 dimensions into two dimensions, where the purple one is tumor case and the red one is non-tumor. The positive and negative cases represented by squared patch features are almost indivisible in Fig. \ref{tsne:d}, while the cases in Fig. \ref{tsne:b} and Fig. \ref{tsne:c} could be separated well, which are represented by our vertical and horizontal patches.

\subsection{Comparison to Other Methods on Kidney Tumor}
\label{sec:comparison}
We extensively compare Crossbar-Net with the low-level methods of kidney tumor segmentation, the multi-scale 2D-CNN model with squared patches, the 3D patch-based CNN model and the image-based CNN models.

As mentioned in Section \ref{sec:related-work}, the low-level feature methods \cite{lee2017detection, hodgdon2015can, linguraru2009computer,linguraru2011automated,skalski2016kidney} have their own specific goals and are difficult to be applied directly here, so we apply their basic operations to our data set: first extracting the whole kidney area with tumors being included firstly, then calculating features manually, and finally classifying or segmenting tumors with non-CNN methods. These methods are termed as hand-crafted based methods (\textbf{HCM}). Besides, our Crossbar-Net is essentially a patch-based multi-scale CNN model with local and contextual information being considered. Hence, the second and third compared methods are the multi-scale \textbf{2D-CNN} and multi-scale \textbf{3D-CNN}. The 2D-CNN model was modified from \cite{moeskops2016automatic}, in which all parameters are kept except for the nodes of the output layer are changed from 9 to 2. The basic 3D-CNN in \cite{dey2018diagnostic} is adopted as the 3D model, in which the input 3D patches are $50 \times 50 \times 5$ and $100 \times 100 \times 10$ and the size of the convolutional kernels is $3 \times 3 \times 3$. The patches are extracted with \emph{basic sampling strategy} in 2D-CNN and 3D-CNN models.

As for the image-based CNN methods, we investigate four models which are representative in medical image segmentation: the \textbf{FCN-G} \cite{zhang2017combining}, the \textbf{U-Net} \cite{ronneberger2015u}, the \textbf{V-Net} \cite{milletari2016v} and the \textbf{SegCaps} \cite{lalonde2018capsules}. The FCN-G is a combination of FCN model and graph model, and the later is a post-process of the former segmentation results. Similar to the original literature \cite{zhang2017combining}, we also adopt VGGnet \cite{simonyan2014very} as basic architecture of the FCN and apply transfer learning on it. The U-Net, a popular image-based CNN model, is a promising model in medical image segmentation. We implement it with Python 3.6.8 \cite{van2009python} and PyTorch 0.4.0 \cite{paszke2009auto}. The best test results are obtained after 28 epoches in this model. The V-net is another typical image-based model, which is also a 3D segmentation model. The SegCaps is also a representative segmentation model and we get the code from Github \cite{segcaps2018code}. We implemented this model with Tensorflow 1.11.0 \cite{abadi2015tensorflow} on 4 GPUs. The segcaps3 is chosen as network and other parameters are kept to be consistent with the original code. To adjust our dataset to this framework, we modified the code that is relevant to reading and converting images. We fed the original images to U-Net and SegCaps, and fed the cropped images to the FCN-G and V-Net to fit the respective input sizes of the two models.

%table4
\begin{table}
\fontsize{7.5}{10}\selectfont
\setlength{\belowcaptionskip}{-0.3cm}
\caption{Comparison among different methods on kidney tumors.}
\label{table_adnc}
\begin{center}
\renewcommand\arraystretch{1}
\begin{tabular}{c|cccc}
\toprule
&DSC&TPF&HD \tiny{(mm)}&CD \tiny{(mm)}\\
\midrule
HCM&0.686&0.788&25.991&11.231\\
FCN-G \cite{zhang2017combining}&0.736&0.752&20.153&9.372 \\ %
U-Net \cite{ronneberger2015u}&0.838&0.832&13.1&4.510 \\ %
V-Net \cite{milletari2016v}&0.887&0.891&10.341&3.895 \\  %
SegCaps \cite{lalonde2018capsules}&0.879&0.882&10.451&4.132\\  %
2D-CNN \cite{moeskops2016automatic}&0.718&0.709&20.982&9.853 \\  %
3D-CNN \cite{dey2018diagnostic}&0.812&0.820&14.225&6.039 \\ %
Crossbar-Net&\textbf{0.913}&\textbf{0.915}&\textbf{8.891}&\textbf{2.624}\\
\bottomrule
\end{tabular}
\end{center}
%\vspace{-0.6cm}
\end{table}

%figure11
\begin{figure*}[hbtp]
\centering
\includegraphics[width = 6.4in, height=8.5in]{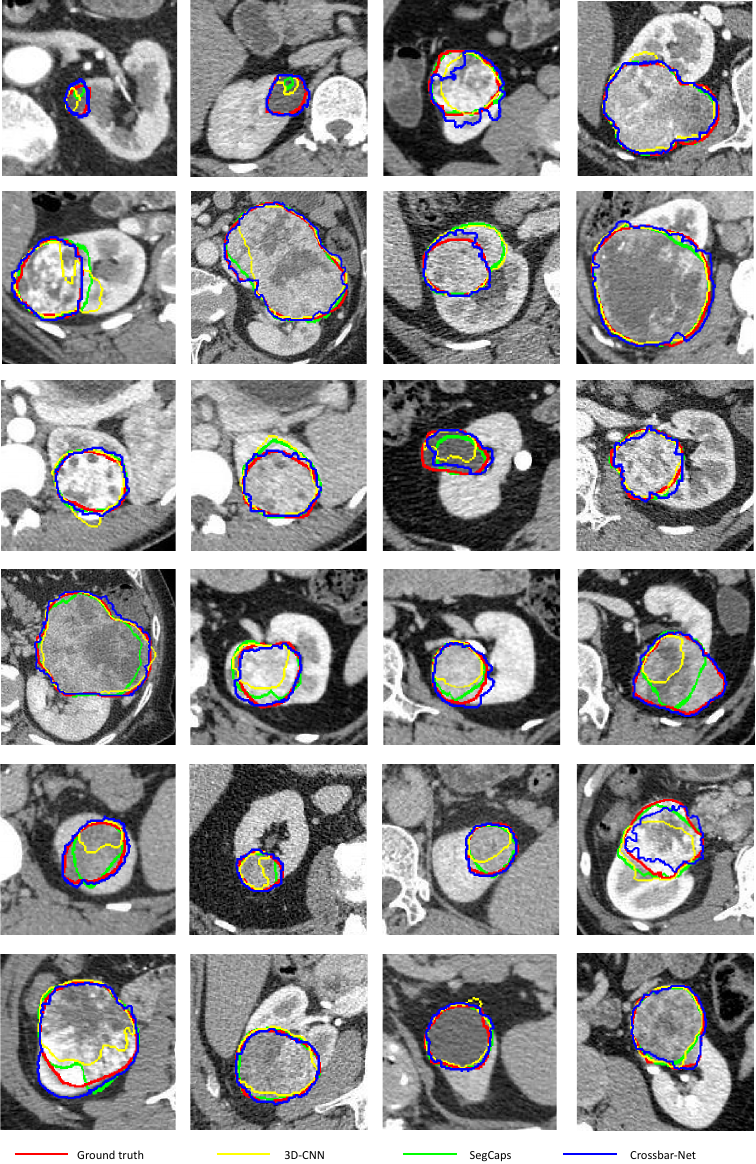}
\caption{Examples of segmentation results with ground truth on kidney tumor dataset. The red, blue, yellow and green curves are manual annotation, Crossbar-net, multi-scale 3D-CNN \cite{dey2018diagnostic}, and SegCaps \cite{lalonde2018capsules} contour, respectively.} \label{kidney_seg_compare}
\end{figure*}

We sampled testing images from one testing set which consists of 30 subjects with totally 600 images being randomly sampled. To fully observe all these 600 tumors, we depicted the kernel density estimation of DSC of all compared methods in Fig. \ref{dr_bar}. It can be observed that Crossbar-Net achieves promising DSC (larger than 0.9) on most tumors. Many low DSC distributed in the multi-scale 2D-CNN and the HCM. The U-Net performs better than these two methods and multi-scale 3D-CNN. SegCaps achieves the second best performance. There are some cases of 0 DSC in the results of U-Net and SegCaps, which indicates that some tumors are missed by these two models.

In Table \ref{table_adnc}, we list average values of DSC, TPF, HD and CD of all test sets of different methods. It is obvious that Crossbar-Net outperforms other methods in terms of the higher DSC and TPF. Moreover, as shown in Table \ref{table_adnc}, it is predominant that Crossbar-Net obtains the smallest value of HD and CD measurements which reflect high quality segmentation. The multi-scale 3D-CNN obtains a higher DSC than the 2D-CNN model since it takes the spatial information into account. Performance of U-Net is slightly better than that of the multi-scale 3D-CNN. In the FCN-G case, the graph model depends on the result of the FCN model\cite{long2015fully} and the performance is not very competitive. Although the SegCaps is a 2D model, it performs competitively with V-Net, both significantly superior to other methods except to our Crossbar-Net. The FCN-G, U-Net and SegCaps and V-Net are all developed from FCN model. In order to explore the reasons why these methods are not very effective, we have also applied FCN directly in our task. The result is not desirable (the DSC is even $< 0.6$) which ignores the local details especially on the boundary of the small tumors. This may be the reason of 0 DSC cases occurring in the U-Net and SegCaps in Fig. \ref{dr_bar}.
%figure stage submodel
\begin{figure}[tb]
\centering
\includegraphics[width = 3.4in]{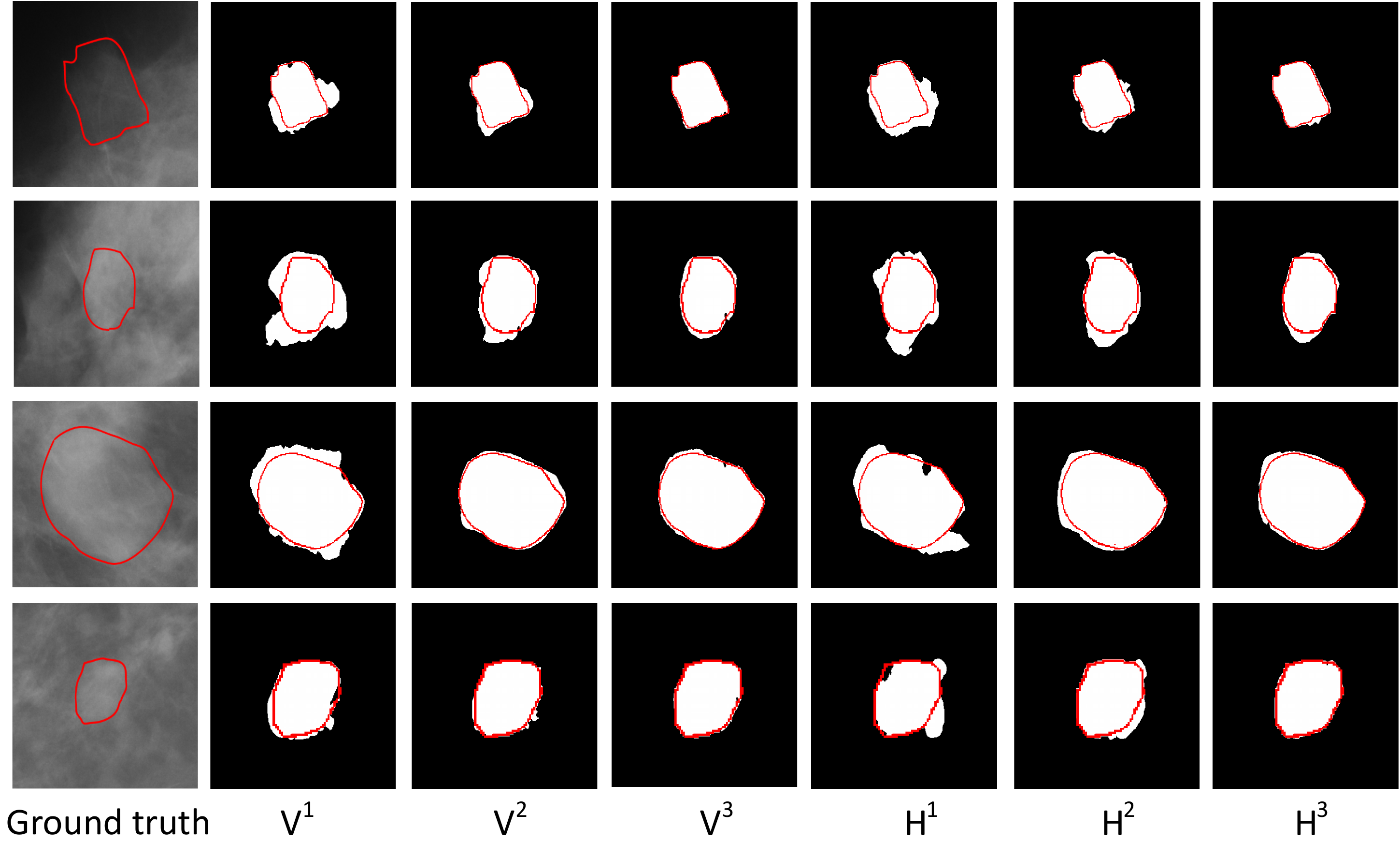}
\caption{Performance of each sub-model on DDSM. The left column is ground truth image, the second to forth columns are three vertical sub-models, and the last three columns are horizontal sub-models, respectively.}\label{DDSM_stage_example}
\end{figure}

\begin{table}
\fontsize{7.5}{8}\selectfont
\setlength{\belowcaptionskip}{-0.2cm}
\caption{Average DSC of each sub-model in LV segmentation and breast mass segmentation.}
\label{average_metrics of heart and breast}
\begin{center}
\renewcommand\arraystretch{1}
\begin{tabular}{p{0.56in}|p{0.27in}p{0.27in}p{0.27in}p{0.27in}p{0.27in}c}
\toprule
  &$\mathbf{V}^1$ &$\mathbf{V}^2$ &$\mathbf{V}^3$ &$\mathbf{H}^1$ &$\mathbf{H}^2$ &$\mathbf{H}^3$\\
\midrule
Breast mass&0.853 &0.875 &0.902&0.849&0.872&0.897\\
LV &0.875&0.881&0.903&0.869&0.883&0.908\\
\bottomrule
\end{tabular}
\end{center}
%\vspace{-0.5cm}
\end{table}

%breast
\begin{table}
\fontsize{7.5}{8}\selectfont
\setlength{\belowcaptionskip}{-0.2cm}
\caption{DSC of each method in breast mass segmentation.}
\label{breast_result_metric}
\begin{center}
\renewcommand\arraystretch{1}
\begin{tabular}{p{0.45in}p{0.55in}p{0.45in}p{0.45in}p{0.55in}}
\toprule
Method\cite{dhungel2015deep}&Cross-sensor\cite{cardoso2017mass}&AM-FCN \cite{zhu2018Adversarial}&Method \cite{zhang2018photoacoustic} &Crossbar-Net\\
\midrule
0.8700&0.9000&0.9130&0.9118&\textbf{0.9122}\\
\bottomrule
\end{tabular}
\end{center}
%\vspace{-0.5cm}
\end{table}
%figure breast
\begin{figure}[tb]
\centering
\includegraphics[width =3.1in]{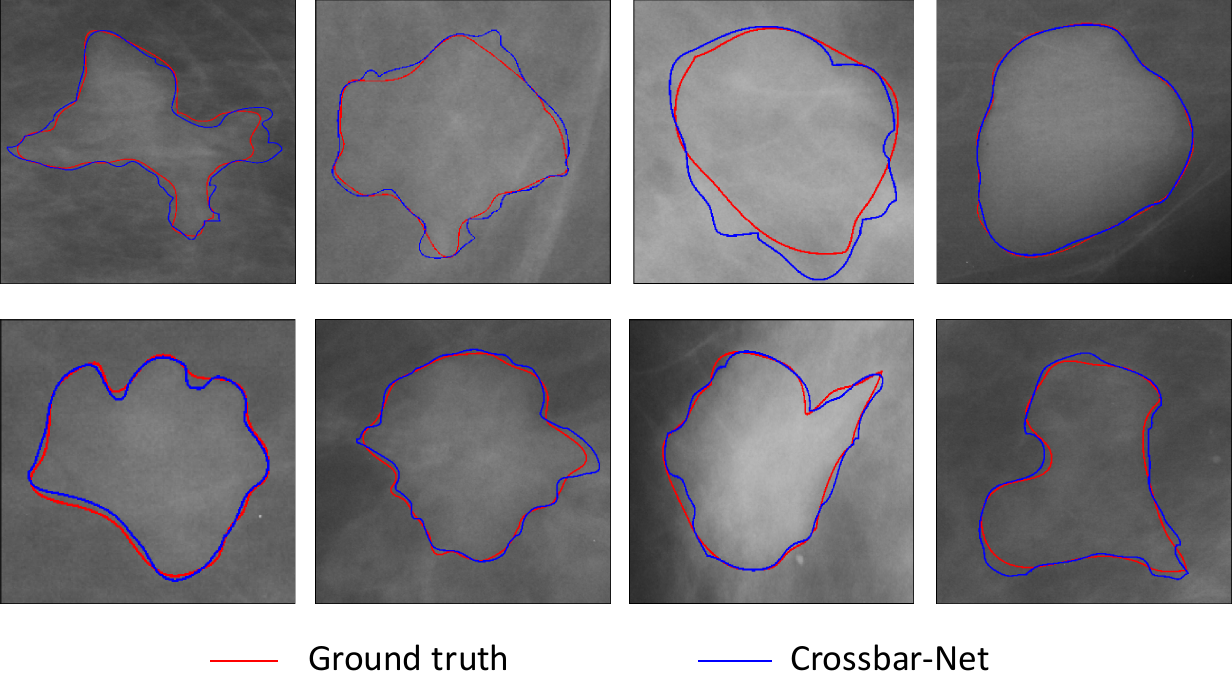}
\caption{Typical segmentation results on DDSM dataset.}\label{mm_segment}
\end{figure}
As shown in Fig. \ref{kidney_seg_compare}, we illustrate several typical segmentation examples of Crossbar-Net, image-based method and 3D patch-based model. Obviously, Crossbar-Net segmentation is similar to the ground truth in most cases. SegCaps performs well in big tumors while fails in small cases, even though the tumor has a distinctive texture. The first and second image in the first row of Fig. \ref{kidney_seg_compare} are the unsatisfactory cases of the SegCaps. The multi-scale 3D-CNN is competitive with SegCaps especially on small tumors, which may be related to its $3 \times 3 \times 3$ small convolutional kernels.

In addition, we have recorded the computation cost of Crossbar-Net, U-Net, and SegCaps: with the implementation on one GPU, our model (including six sub-models) takes $\sim$ 1h for training and less than 1.5s for segmenting a new patient (about 35 slices). The U-Net is very close to our method in training and testing time. The SegCaps running on 4 GPUs takes about 110 minutes for training one epoch and reaches to convergence after 24 epoches. Thanks to our sampling and boosting-like-training, many correctly segmented patches will not feed into the later rounds, which helps reduce the training patches and training time. Specifically, about 580,000 crossbar patches (\emph{i.e.}, 580,000 vertical patches and 580,000 horizontal patches) are input to the vertical and horizontal sub-model respectively in the 1-st training round, to obtain the corresponding $H^{1}$ and $V^{1}$. In the 2-nd round, about 150,000 patches are fed to $H^{1}$ and $V^{1}$ respectively, including $\sim$ 90,000 re-sampling patches and 60,000 basic sampling patches. In the 3-rd round, about 70,000 patches are totally sampled.
%%%%%%%%%%%%%%%%%%%%%%%%%%%%%%%%%%%%%%%%%%%%%%%%%%%%%%%

\subsection{Crossbar-Net for Breast Mass Segmentation}
\label{subsec:breast}
We segment the breast mass in mammograph for evaluating the generalization to the related tasks. We illustrate the performance of each sub-models on DDSM in Fig. \ref{DDSM_stage_example} and Table \ref{average_metrics of heart and breast}, confirming the characteristics of self-improvement and mutual help again. In Table \ref{breast_result_metric}, we list the DSC of Crossbar-Net and several state-of-the-art methods which are implemented on DDSM dataset. In this table, we report the best records of \cite{zhang2018photoacoustic, zhu2018Adversarial, cardoso2017mass,dhungel2015deep} as reported in the original manuscripts. The results in Table \ref{breast_result_metric} demonstrate that Crossbar-Net is slightly superior to others. As shown in Fig. \ref{breastcropping:b}, it is possible that the black filled in some crossbar patches (especially the non-tumor patches) contributes to improving the discrimination of the patches. We also show several segmented visualization results of Crossbar-Net (Fig. \ref{mm_segment}).

Noted that the cost of training and testing on this dataset is larger than that on the kidney and cardiac datasets for the large patch of mammography. This is because that under the $500 \times 50$ and $50 \times 500$ patch size, the structure of the vertical and horizontal sub-models consists of 11 layers of convolution, 4 layers of pooling and 1 softmax layer, respectively. Also, about 500,000 patches are extracted to train $\mathbf{V}^1$ and $\mathbf{H}^1$. Meanwhile, about 120,000 and 50,000 patches are fed to the sub-models in the 2-nd and 3-rd round, respectively. Thus, regarding the more parameters in the segmentation model compared with the kidney tumor datasets, all six sub-models take about 6h for training and about 15s for segmenting a new subject.

 %figure stage submodel
\begin{figure}[tb]
\centering
\includegraphics[width = 3.4in]{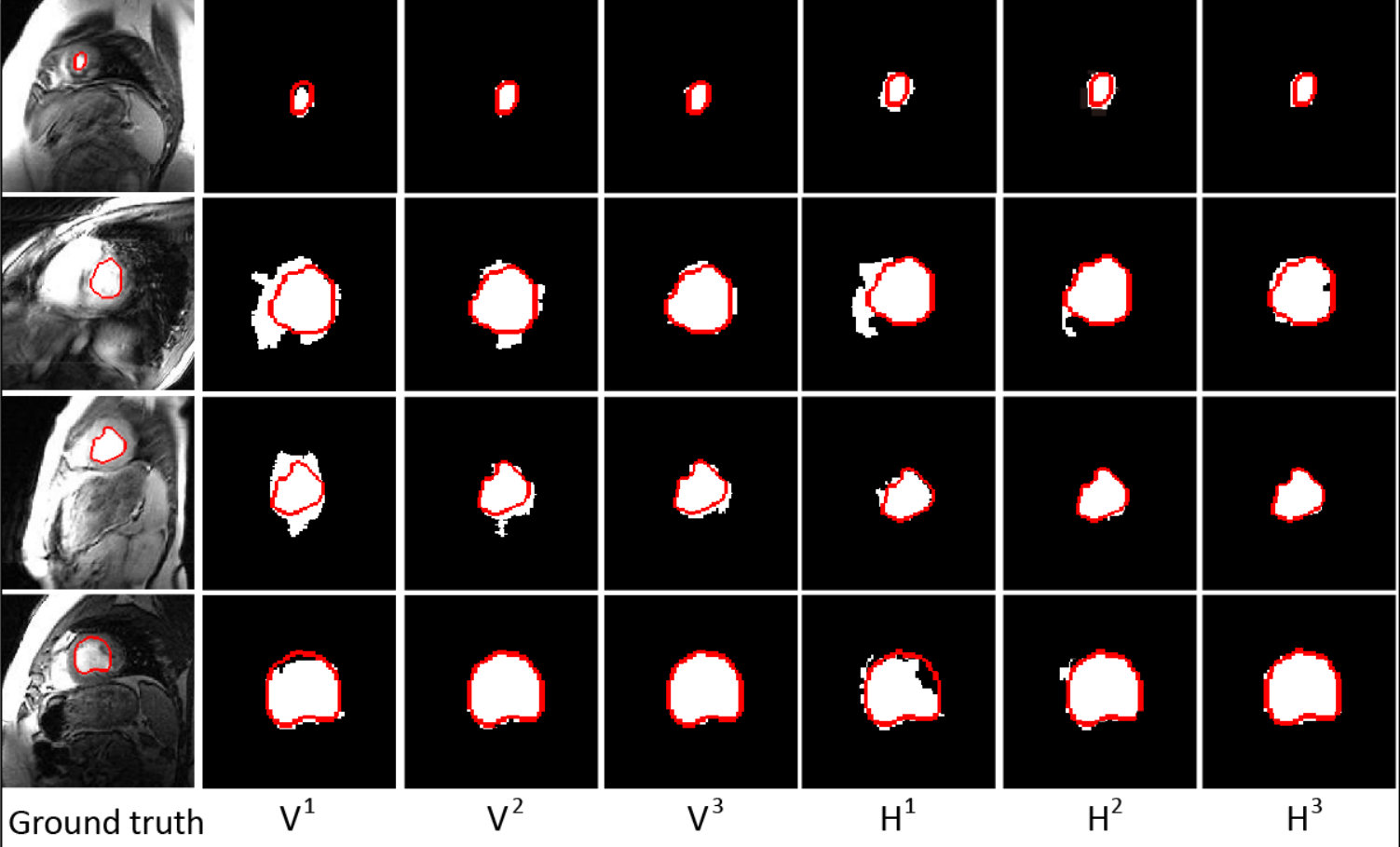}
\caption{Performance of each sub-model in LV cave segmentation on the cardiac dataset. The left column is ground truth image, the second to forth columns are three vertical sub-models, and the last three columns are horizontal sub-models, respectively.}\label{heart_stage_example}
\end{figure}

\begin{table}[tp]
  \centering
  \renewcommand\arraystretch{1.2}
  \fontsize{7.5}{8}\selectfont
 % \begin{threeparttable}
  \caption{Prediction performance comparison in cardiac segmentation.}
  \label{tab:performance_comparison}
    \begin{tabular}{cccccc}
    \toprule
    \multirow{2}{*}{Method}&
    \multicolumn{3}{c}{LV }&\multicolumn{2}{c}{MYO}\\
    \cmidrule(lr){2-4} \cmidrule(lr){5-6}
    &DSC&AD \tiny{(mm)} & HD \tiny{(mm)} & DSC & HD \tiny{(mm)}\\
    \midrule
    DMWDP \cite{santiago2017fast}&0.859&2.10&-&-&-\\
    ASAMM \cite{santiago2018combining}&0.856&2.30&-&-&-\\
    DC-FCN \cite{khened2018densely}&0.915&-&12.08&0.855&14.98\\
    3D-CNN \cite{patravali20172d-3d}&0.925&-&14.65&0.855&38.12\\
    GridNet \cite{zotti2017gridnet}&\textbf{0.955}&-&5.85&0.885&8.01\\
    Crossbar-Net&0.925&\textbf{1.82}&\textbf{3.60}&\textbf{0.892}&\textbf{4.63}\\
    \bottomrule
    \end{tabular}
   % \end{threeparttable}
\end{table}

%figure cardic
\begin{figure*}[tb]
\vspace{-0.8cm}
\setlength{\abovecaptionskip}{-0.2cm}
\centering
\includegraphics[width = 7.2in]{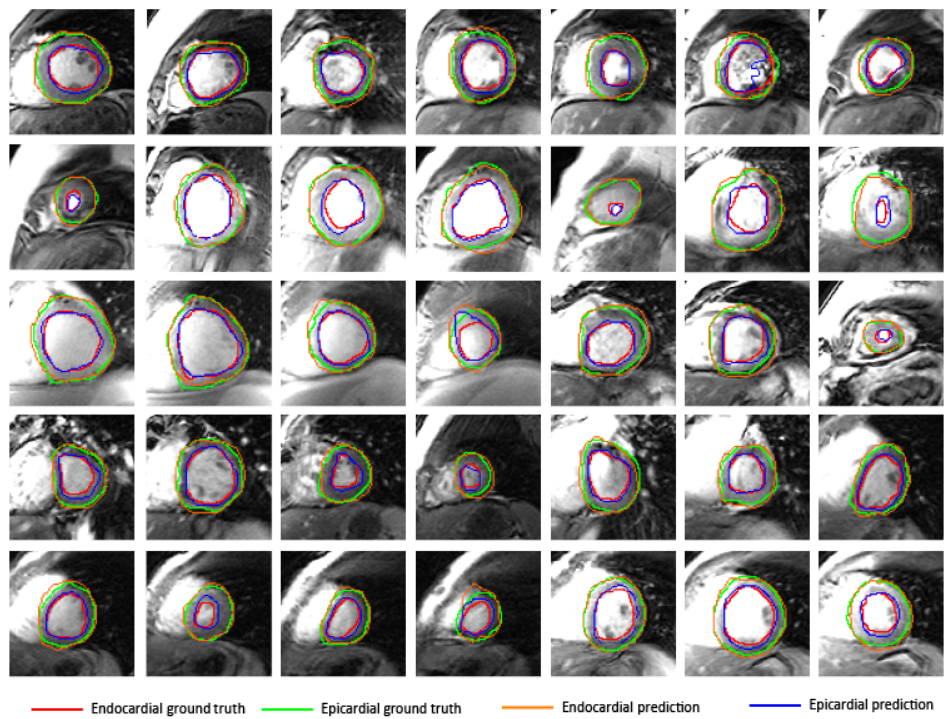}
\caption{Examples of resulting segmentations for subjects 24-33 of the cardiac dataset \cite{andreopoulos2008efficient}. }\label{heart_seg_result}
\end{figure*}
 \subsection{Crossbar-Net for Cardiac Segmentation}
\label{subsec:cardiac}
In this experiment, the LV and myocardium (MYO) in cardiac MRI dataset are segmented. MYO is determined by endocardia and epicardium together, while LV is determined by the endocardia. The performances of sub-models in LV segmentation are visualized in Fig. \ref{heart_stage_example} and quantified in Table \ref{average_metrics of heart and breast}. Both the figure and the table indicate a gradual improvement in performance among sub-models in the same direction of different rounds.

Also, we report the DSC and distance metrics of our method and the other five methods proposed in \cite{santiago2017fast, santiago2018combining, khened2018densely, patravali20172d-3d,zotti2017gridnet} in Table \ref{tab:performance_comparison}. The deformable model (DMWDP) in \cite{santiago2017fast} and Active Shape and Motion Model (ASAMM) in \cite{santiago2018combining} are non-CNN models, both of which are applied on the same dataset with Crossbar-Net. The remaining three methods are CNN-based models applied on other dataset. All results are directly reported from their original articles for comparison. The average perpendicular distance (\textbf{AD}) in Table \ref{tab:performance_comparison} corresponds to the average distance between each pixel in the predicted boundary and the closest ground truth pixel. It is observed that DSC of Crossbar-Net in LV segmentation stands the second best overall result, with GridNet \cite{zotti2017gridnet} being the most accurate. However, our HD and AD are superior to that of other methods. For the MYO segmentation, our method outperforms all other methods. The highest DSC of MYO and slightly low DSC of LV means that Crossbar-Net is superior to other methods in epicardium segmentation. Therefore, our method is competitive compared with the state-of-the-art methods in cardiac segmentation.

We also show some of the segmented visualization results of our method. As shown in Fig. \ref{heart_seg_result}, several representative samples from each sequence of subject 24-33 are illustrated. Note that the performance of Crossbar-Net is better in cardiac segmentation than on kidney tumors because the shape of the LV cavity and the myocardium are more regular.

\section{Conclusion}
\label{sec:conclusion}
In this paper, we propose a novel segmentation model named as Crossbar-Net, in which the innovations focus on the shape of patches, the way of patch sampling and the style of cascaded training. For the shape of patches, the crossbar patches cover the kidney tumor in both horizontal and vertical directions and capture the local and contextual information simultaneously. For the way of sampling patches, the \emph{basic sampling strategy} and \emph{covering re-sampling strategy} are proposed. The combination of these two strategies not only enhances the role of mis-segmented regions but also prevents sub-models from being over-emphasized on the mis-segmented regions. For the cascaded training style, the segmentation result of sub-models in one direction can be complemented by sub-models in the other direction, and each sub-model can perform self-improvement with re-sampling the mis-segmented region. Our model can simultaneously learn a variety of information and achieve promising segmentation results on different size, shape, contrast and appearance of kidney tumors. Moreover, the successful application on cardiac and breast mass segmentation shows that Crossbar-Net has a wide range of application. The future work is to extend the direction of symmetric information from horizontal and vertical axes to the other axes.
%\appendices
%\section{Proof of the Training Error Bound}
%\label{proof_app}
%\section{}
%% use section* for acknowledgment
%\ifCLASSOPTIONcompsoc
%  % The Computer Society usually uses the plural form
%\else
%  % regular IEEE prefers the singular form
%\fi

\ifCLASSOPTIONcaptionsoff
  \newpage
\fi



%\bibliographystyle{IEEEtran}

\bibliographystyle{IEEEtran}
\bibliography{egpaper_final}

\balance

\end{document}